\documentclass[twoside]{article}

\usepackage[accepted]{aistats2025}

\usepackage{latexsym}
\usepackage{amssymb}
\usepackage{amsmath}
\usepackage{amsthm}
\usepackage{booktabs}
\usepackage{enumitem}
\usepackage{graphicx}
\usepackage{color}
\usepackage{xcolor,colortbl}
\usepackage{url}
\usepackage{multirow}
\usepackage{mathtools} % amsmath with fixes and additions
\usepackage{booktabs} % commands to create good-looking tables
\usepackage{amsfonts}
\usepackage{subfig}
\usepackage{tikz} % nice language for creating drawings and diagrams
\usepackage{algorithm,algorithmic}
\usepackage{amsthm}
\usepackage{amsfonts}
\usepackage{amssymb}
\usepackage{amsopn}
\usepackage{amsmath}
\usepackage{bbm}

\newlength{\Oldarrayrulewidth}

\definecolor{DarkGray}{gray}{0.6}
\definecolor{Gray}{gray}{0.85}
\definecolor{LightGray}{gray}{0.95}
\definecolor{LightCyan}{rgb}{0.88,1,1}
\newcolumntype{a}{>{\columncolor{DarkGray}}l}
\newcolumntype{b}{>{\columncolor{Gray}}l}
\usepackage[round]{natbib}

\newcommand{\STAB}[1]{\begin{tabular}{@{}c@{}}#1\end{tabular}}

\begin{document}



\twocolumn[

\aistatstitle{Bayesian Inference in Recurrent Explicit Duration Switching Linear Dynamical Systems}

\aistatsauthor{ Mikołaj Słupiński \And Piotr Lipiński }

\aistatsaddress{ Computational Intelligence Research Group,\\Institute of Computer Science, University of Wrocław,\\ Wrocław, Poland\\ \texttt{\{mikolaj.slupinski,piotr.lipinski\}@cs.uni.wroc.pl}} ]

\begin{abstract}
  In this paper, we propose a novel model called Recurrent Explicit Duration Switching Linear Dynamical Systems (REDSLDS) that incorporates recurrent explicit duration variables into the rSLDS model. We also propose an inference and learning scheme that involves the use of Pólya-gamma augmentation. We demonstrate the improved segmentation capabilities of our model on three benchmark datasets, including two quantitative datasets and one qualitative dataset.
\end{abstract}

\section{Introduction}
Understanding complex dynamics of spatio-temporal data is a challenging task studied with a range of techniques, from theoretical mathematical models based on partial differential equations to practical machine learning ones constructed on recorded data. One of such techniques are State Space Models (SSMs) \citep{murphy_state-space_2023}, including Hidden Markov Models (HMMs), Switching Linear Dynamical Systems (SLDS) \citep{fox_bayesian_2011}, and their numerous extensions. SSMs try to describe the complex dynamics by decomposing it into some separate parts with much simpler dynamics, corresponding to some hidden states of the complex system, and a transition mechanism between such hidden states.

In this paper, we focus on Recurrent Switching Linear Dynamical Systems (rSLDS) \citep{linderman_bayesian_2017} which are one of the popular modern extensions of SLDS, incorporating some machine learning concepts to model the dynamics more accurately. SLDS assume that each observable temporal value is strongly related to an unobservable discrete hidden state and an unobservable value of a latent continuous variable (sometimes referred to as a continuous hidden state) that determine the dynamics of the complex system. They model the recorded temporal values as observations of a certain random variable, and the probabilistic distribution of the random variable depends on the current discrete hidden state and the current value of the latent continuous variable. Discrete hidden states are also modeled as (unrecorded) observations of a certain (latent) random variable, and the probabilistic distribution of the random variable is usually defined by a matrix of probabilities of transitions from one discrete hidden state to another. Similarly, the (unrecorded) values of the latent continuous variable are modeled as observations of a certain (latent) random variable that depends on the current discrete hidden state. rSLDS assume that the next discrete hidden state also depends on the previous continuous hidden states.

To illustrate the concept of SLDS, we may consider a simple example of modeling the movement of a person. Discrete hidden states correspond to different modes of transport. Continuous hidden states correspond to the true positions of the person on the map. The recorded temporal values correspond to the observed (noised) positions of the person in successive time instants.
\begin{figure}[t]
\centering
\includegraphics[width=0.8\linewidth]{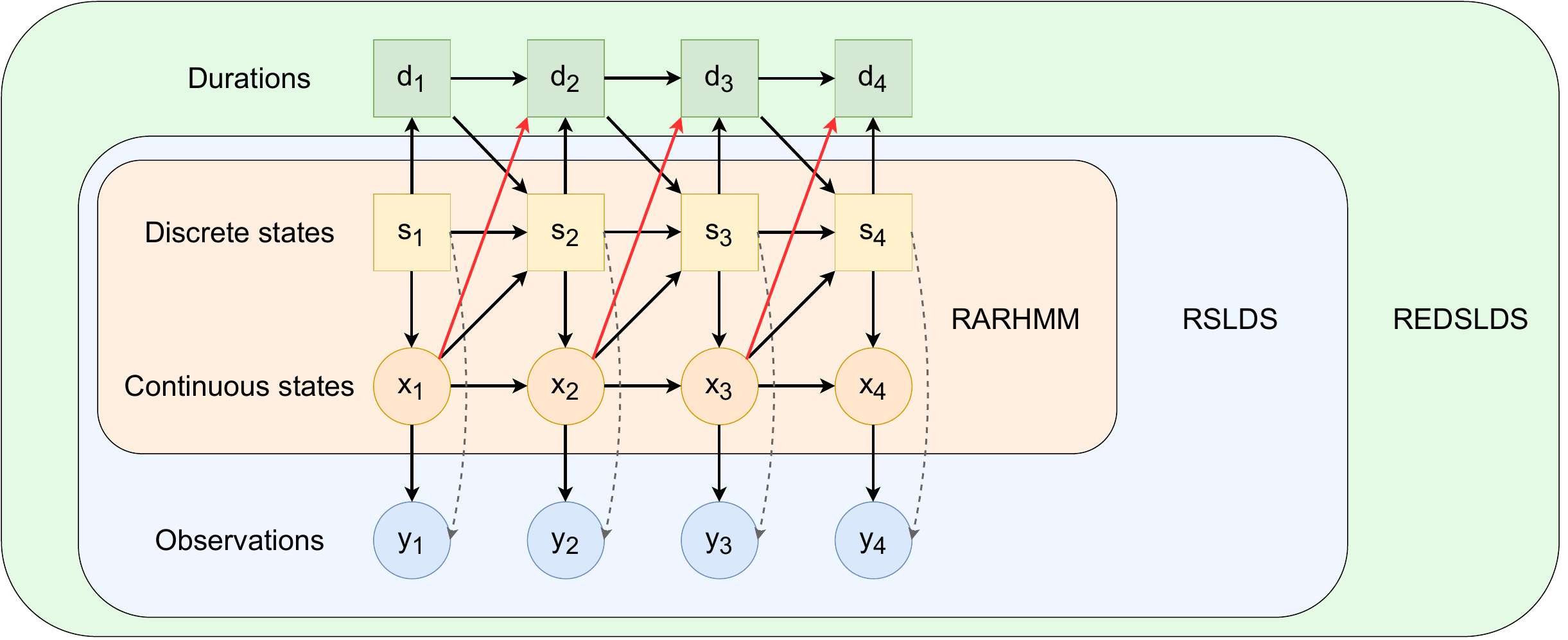}
\caption{The graphical model of REDSLDS}
\label{fig:graphical_model}
\end{figure}
Due to their interpretability, (r)SLDS models  are widely used for neural activity analysis \citep{zoltowski_general_2020, osuna-orozco_identification_2023, song_unsupervised_2023, jiang_dynamic_2024, bush_latent_2024} and more recently for vehicle trajectory prediction \citep{wei_navigation_2024}.

In regular SLDS, switching between particular discrete hidden states is state-dependent only, since the probability of transition to the next state depends only on the previous hidden state. In rSLDS, switching is also state-dependent only, but the probability of transition to the next state depends also on the previous continuous state. Many studies \citep{dewar_inference_2012,chiappa_explicit-duration_2014,ansari_deep_2021} suggested that, in the case of more complex systems, the use of state-dependent switching only makes it difficult to capture the complex dynamics in the learning process without further limiting the duration of the discrete hidden state. Therefore, a type of time-dependent switching was incorporated in SSMs such as hidden Markov models with implicit state duration distributions \citep{dewar_inference_2012}, implicit-duration Markov switching models \citep{chiappa_explicit-duration_2014}, or deep implicit-duration switching model \citep{ansari_deep_2021}.

As stated by \citet{gelman_bayesian_2020}, there exist three distinct approaches to consider when it comes to modeling and prediction.

From a conventional statistical viewpoint, textbooks usually present statistical inference as a pre-defined problem where a model has been pre-selected. In the context of Bayesian statistics, the aim is to provide an accurate summary of the posterior distribution. The computation process should continue until an approximate convergence is achieved.

From the viewpoint of machine learning, the primary aim in machine learning is typically prediction, rather than estimating parameters, and the computational process can be halted once the accuracy of cross-validation prediction has reached a steady state.

From the viewpoint of model exploration, a substantial portion of our work in applied statistics is dedicated to probing, experimenting with a range of models, a significant number of which may exhibit poor alignment with data, subpar predictive capabilities, and sluggish convergence.

Our approach aligns with the perspective of model exploration, as iterative model building has become standard in Bayesian data analysis \citep{gelman_bayesian_2020, gelman_bayesian_2015, blei_build_2014, moran_holdout_2024}.

This paper proposes an extension of rSLDS with an additional mechanism to control the duration of the discrete hidden state. It introduces a recurrent explicit state duration variable that is modeled using a conditionally categorical random variable with a given finite support, similarly to explicit duration SSMs, such as HMMs \citep{yu_hidden_2016} and other switching dynamical systems with continuous states \citep{chiappa_explicit-duration_2014,chiappa_movement_2010,oh_learning_2008}. Such an explicit duration mechanism
facilitates the learning process and makes it easier to capture the complex dynamics that outperforms the original rSLDS.

The main contributions of this work are: REDSLDS -  a probabilistic model using locally linear dynamics that jointly
approximate global nonlinear dynamics, incorporating recurrent explicit duration variables for more accurate segmentation, a fully-Bayesian sampling procedure using Pólya-Gamma data augmentation to allow for fast and
conjugate Gibbs sampling, a series of experiments on synthetic and real popular datasets, assesing our model's performance on variety of problems.

\section{Background}
Our research is based on the findings in the fields of switching state-space models and explicit duration modeling.
\subsection{Switching State-Space Models}
There are numerous models available that utilize both state-space modeling and discrete Markovian states. However, it can be argued that Switching Linear Dynamical Systems serve as the basic foundation for more advanced constructions.

The Switching Linear Dynamical System \citep{kim_dynamic_1994, murphy_switching_1998} is the generative model, defined as follows. At each time $t=1,2, \ldots, T$ there is a discrete latent state $s_{t} \in\{1,2, \ldots, K\}$ following Markovian dynamics,
\begin{equation}
    s_{t+1} \mid s_{t},\left\{\boldsymbol{\pi}_{k}\right\}_{k=1}^{K} \sim \boldsymbol{\pi}_{s_{t}},
\end{equation}
where $\left\{\boldsymbol{\pi}_{k}\right\}_{k=1}^{K}$ is the Markov transition matrix and $\boldsymbol{\pi}_{k}\in[0,1]^{K}$ is its $k$-th row. In addition, a continuous latent state $\mathbf{x}_{t} \in \mathbb{R}^{M}$, where $M$ is the dimensionality of the continuous latent space, follows conditionally linear dynamics, where the discrete state $s_{t}$ determines the linear dynamical system used at time $t$:
\begin{equation}
\mathbf{x}_{t+1}=\mathbf{A}_{s_{t+1}} \mathbf{x}_{t} + \mathbf{a}_{s_{t+1}} + \mathbf{e}_{t}, \quad \mathbf{e}_{t} \stackrel{\text { iid }}{\sim} \mathcal{N}\left(0, \mathbf{Q}_{s_{t+1}}\right),
\end{equation}
for matrices $\mathbf{A}_{k}, \mathbf{Q}_{k} \in \mathbb{R}^{M \times M}$ and bias vector $\mathbf{a}_{k}, \in \mathbb{R}^{M}$ (where $\mathbf{A}_{k}$ is linear transformation matrix between subsequent continuous states, and $\mathbf{Q}_{k}$ is their covariance) for $k=1,2, \ldots, K$. Finally, at each time $t$ a linear Gaussian observation $\mathbf{y}_{t} \in \mathbb{R}^{N}$, where $N$ is the dimensionality of the observation space, is generated from the corresponding latent continuous state,
\begin{equation}
\mathbf{y}_{t}=\mathbf{C}_{s_{t}} \mathbf{x}_{t}+ \mathbf{c}_{s_{t+1}}+\mathbf{f}_{t}, \quad \mathbf{f}_{t} \stackrel{\mathrm{iid}}{\sim} \mathcal{N}\left(0, \mathbf{S}_{s_{t}}\right),
\end{equation}
for $\mathbf{C}_{k} \in \mathbb{R}^{N \times M}, \mathbf{c}_{k} \in \mathbb{R}^{N}, \mathbf{S}_{k} \in \mathbb{R}^{N \times N}$, where $\mathbf{C}_{k}$ is an emission matrix, $\mathbf{c}_{k}$ is an emission bias vector, and $\mathbf{S}_k$ is covariance of the measurement noise. The system parameters comprise the discrete Markov transition matrix and the library of linear dynamical system matrices, which we write as
\begin{equation}
\theta=\left\{\left(\boldsymbol{\pi}_{k}, \mathbf{A}_{k}, \mathbf{a}_{k}, \mathbf{Q}_{k}, \mathbf{C}_{k}, \mathbf{c}_{k}, \mathbf{S}_{k}\right)\right\}_{k=1}^{K} .
\end{equation}
% For simplicity, we will require $C, S$, and $d$ to be shared among all discrete states in our experiments.

To learn an SLDS using Bayesian inference \citep{fox_bayesian_2011}, we place conjugate Dirichlet ($\operatorname{Dir}$) priors on each row of the transition matrix and conjugate matrix normal inverse Wishart (MNIW) priors on the linear dynamical system parameters, writing
\begin{equation}
\begin{gathered}
\boldsymbol{\pi}_{k}\left|\alpha \stackrel{\mathrm{iid}}{\sim} \operatorname{Dir}(\alpha), \quad (\mathbf{A}_{k}, \mathbf{a}_{k}), \mathbf{Q}_{k}\right| \lambda \stackrel{\mathrm{iid}}{\sim} \operatorname{MNIW}(\lambda), \\
(\mathbf{C}_{k}, \mathbf{c}_{k}), \mathbf{S}_{k} \mid \eta \stackrel{\mathrm{iid}}{\sim} \operatorname{MNIW}(\eta),
\end{gathered}
\end{equation}
where $\alpha, \lambda$, and $\eta$ denote hyperparameters.

\subsection{Pólya-gamma Augmentation}

An approach that has previously been used to improve the capabilities of SLDS \citep{linderman_bayesian_2017,nassar_tree-structured_2019} involved incorporating a change in the probabilities of the switch based on the latent state. In other words, our goal is to have $p(s_t \mid s_{t-1},x_{t-1}) \neq p(s_t \mid s_{t-1})$. To achieve this, a mapping from continuous to discrete space is required. An approach that enables such a mapping is known as Pólya-gamma augmentation.
The main result in \citep{polson_bayesian_2013} shows that binomial probabilities can be expressed as a combination of Gaussians using a Pólya-gamma distribution. The key discovery is based on the integral identity, where $b>0$, given by
\begin{equation}
\frac{\left(e^\psi\right)^a}{\left(1+e^\psi\right)^b}=2^{-b} e^{\kappa \psi} \int_0^{\infty} e^{-\omega \psi^2 / 2} p(\omega) d \omega,
\end{equation}
where $\kappa=a-b / 2$ and $\omega$ is a random variable whose pdf $p_{\mathrm{PG}}(\omega \mid b, 0)$ is the density of the Pólya-gamma distribution, $\mathrm{PG}(b, 0)$, which does not depend on $v$, and $\psi~\in~\mathbb{R}$.
This technique entails sampling from a Gaussian distribution for the main parameters and a Pólya-gamma distribution for the latent variables.

Suppose the observation of the system at time $t$ follows
\begin{equation}\label{eq:logisticregression}
\begin{gathered}
p\left(w_{t+1} \mid \mathbf{x}_t \right)=\operatorname{Bern}\left(\sigma\left(v_t\right)\right)=\frac{\left(e^{v_t}\right)^{w_{t+1}}}{1+e^{v_t}}, \\
v_t=\mathbf{R}^T \mathbf{x}_t +r,
\end{gathered}    
\end{equation}

where $\operatorname{Bern}$ is Bernoulli's distribution pmf, $\sigma$ is the logistic function, $\mathbf{R} \in \mathbb{R}^{d_x}, r \in \mathbb{R}$.
Then if we introduce the PG auxiliary variables $\omega_{k, t}$, conditioning on $\omega_{1: T}$, \eqref{eq:logisticregression} becomes
$$
\begin{aligned}
p\left(w_t \mid x_t, \omega_{t}\right) & =e^{-\frac{1}{2}\left(\omega_{t} v_{t}-2 \kappa_{t} v_{t}\right)} \\
& \propto \mathcal{N}\left(\mathbf{R}^T \mathbf{x}_t+r \mid \kappa_{t} / \omega_{t}, 1 / \omega_{t}\right)
\end{aligned}
$$
where  $\kappa_{t}=w_{t}-\frac{1}{2}$.
\subsection{Recurrent Models}
SLDS serve as the basis for more sophisticated models, which have been investigated in recent years.
\citet{linderman_bayesian_2017} proposed novel methods for learning Recurrent Switching Linear Dynamical Systems (rSLDS).

They used information on position in the latent hyperplane to influence the probability of future hidden states, thus \emph{incorporating non-stationarity} into the state distribution (namely $p(s_t|s_{t-1},\mathbf{x}_{t-1}) \neq p(s_t|s_{t-1})$)).

\begin{figure}[htbp]
\includegraphics[width=0.45\textwidth]{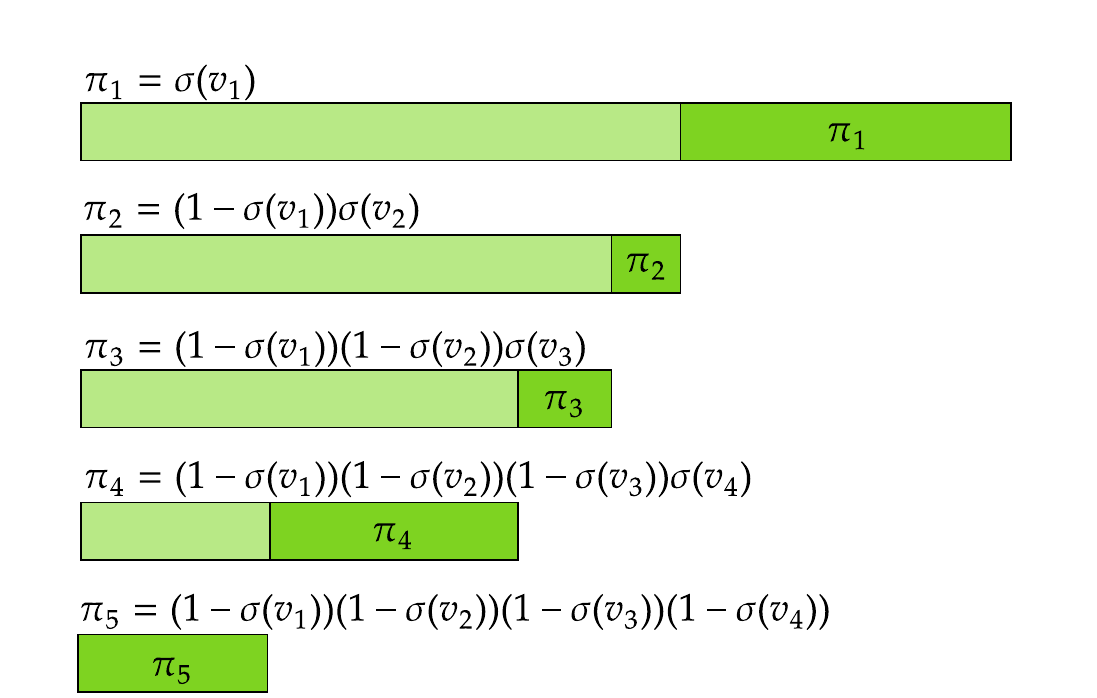}
\caption{The visualization of stick-breaking.}
\label{fig:sb}
\end{figure}
% {\color{blue}There are many variations of incorporating explicit duration variables into both} Switching Linear Dynamical Systems \cite{oh_learning_2008,chiappa_explicit-duration_2014,chiappa_movement_2010} or Switching Nonlinear Dynamical Systems \cite{ansari_deep_2021}.

Another component of recurrent SLDS is stick-breaking logistic regression, and for \emph{efficient block inference updates}, we use a recent Pólya-gamma augmentation strategy \citep{linderman_bayesian_2017}. This augmentation allows certain logistic regression evidence potentials to appear as \emph{conditionally Gaussian} potentials in an augmented distribution, enabling our fast inference algorithms.

% Consider a logistic regression model from regressors $x \in \mathbb{R}^M$, {\color{blue}where $M$ is the dimensionality of the latent space,} to a categorical distribution on the discrete variable $s \in\{1,2, \ldots, K\}$, written as
Consider a logistic regression model from the regressors $\mathbf{x} \in \mathbb{R}^M$ to a categorical distribution on the discrete variable $s \in\{1,2, \ldots, K\}$, written as
\begin{equation}
    s \mid \boldsymbol{x} \sim \pi_{\mathrm{SB}}(\boldsymbol{v}^{S}), \quad v^{S}=\mathbf{R}^{S} \mathbf{x}+\boldsymbol{r},
\end{equation}
where $\mathbf{R} \in \mathbb{R}^{(K-1) \times M}$ is a weight matrix and $\boldsymbol{r} \in \mathbb{R}^{K-1}$ is a bias vector. Unlike standard multiclass logistic regression, which uses a softmax link function, we instead use a stick-breaking link function $\pi_{\mathrm{SB}}: \mathbb{R}^{K-1} \rightarrow[0,1]^K$ (see Figure \ref{fig:sb}), which maps a real vector to a normalized probability vector via the stick-breaking process.
\begin{equation}
    \begin{gathered}
\pi_{\mathrm{SB}}(\boldsymbol{v}^{S})=\left(\begin{array}{lll}
\pi_{\mathrm{SB}}^{(1)}(\boldsymbol{v}^{S}) & \cdots & \pi_{\mathrm{SB}}^{(K)}(\boldsymbol{v}^{S})
\end{array}\right) \\
\pi_{\mathrm{SB}}^{(k)}(\boldsymbol{v}^{S})=\sigma\left(\boldsymbol{v}^{S}_k\right) \prod_{j<k}\left(1-\sigma\left(\boldsymbol{v}^{S}_j\right)\right) \\
=\sigma\left(\boldsymbol{v}^{S}_k\right) \prod_{j<k} \sigma\left(-\boldsymbol{v}^{S}_j\right)
\end{gathered}
\end{equation}
for $k=1,2, \ldots, K-1$ and $\pi_{\mathrm{SB}}^{(K)}(\boldsymbol{v}^{S})=\prod_{k=1}^K \sigma\left(-\boldsymbol{v}^{S}_k\right)$, where $\sigma(x)=e^x /\left(1+e^x\right)$ denotes the logistic function. The probability mass function $p(s_t \mid \mathbf{x}_{t-1})$ is
\begin{equation}\label{eq:state}
    p(s_t \mid x_t-1)=\prod_{k=1}^K \sigma\left(\boldsymbol{v}^{S}_k\right)^{\mathbb{I}[s_t=k]} \sigma\left(-\boldsymbol{v}^{S}_k\right)^{\mathrm{I}[s_t>k]},
\end{equation}
where $\mathbb{I}[\cdot]$ denotes an indicator function that takes value 1 when its argument is true and 0 otherwise.
If we use this regression model as a likelihood $p(s \mid \mathbf{x})$ with a Gaussian prior density $p(\mathbf{x})$, the posterior density \emph{$p(\mathbf{x} \mid s)$ is non-Gaussian and does not admit easy Bayesian updating}. However, \citet{linderman_dependent_2015} show how to introduce Pólya-gamma auxiliary variables $\boldsymbol{\omega}=\left\{\omega_k\right\}_{k=1}^K$ so that the conditional density $p(\mathbf{x} \mid s, \boldsymbol{\omega})$ becomes Gaussian. In particular, by choosing $\omega_k \mid \mathbf{x}, s \sim \operatorname{PG}\left(\mathbb{I}[s \geq k], v^{S}_k\right)$, we have,
$$
\mathbf{x} \mid s, \omega \sim \mathcal{N}\left(\mathbf{\Omega}^{-1} \boldsymbol{\kappa}, \mathbf{\Omega}^{-1}\right),
$$
where the mean vector $\mathbf{\Omega}^{-1} \boldsymbol{\kappa}$ and covariance ma$\operatorname{trix} \mathbf{\Omega}^{-1}$ are determined by
$$
\mathbf{\Omega}=\operatorname{diag}(\omega), \quad \kappa_k=\mathbb{I}[s=k]-\frac{1}{2} \mathbb{I}[s \geq k] .
$$
Thus instantiating these auxiliary variables in a Gibbs sampler or variational mean field inference algorithm \emph{enables efficient block updates while preserving the same marginal posterior distribution $p(\mathbf{x} \mid s)$}.

rSLDS was further extended by \citet{nassar_tree-structured_2019} by adding a tree-based structure of recurrent probabilities. One of the most recent extensions of rSLDS was proposed by \citet{wang_bayesian_2024}, who incorporated higher-order dependence into rSLDS.
A similar approach to rSLDS was applied by \citet{dong_collapsed_2020}. However, the authors replaced linear transformations with neural networks. 
\subsection{Explicit Duration Modeling}

One may observe that, using classic Markovian transitions, the duration time of a single state always follows the geometric distribution, and that may not be true for real-life data.

Explicit Duration Switching Dynamical Systems are a family of models that introduce additional random variables to explicitly model the switch duration distribution.
Explicit duration variables have been applied to both HMMs \citep{dewar_inference_2012,yu_hidden_2016} and SDS with continuous states \citep{chiappa_explicit-duration_2014,chiappa_movement_2010,oh_learning_2008}.
Several methods have been proposed in the literature for modeling the duration of the switch, for example, using decreasing or increasing count and duration indicator variables \citep{chiappa_explicit-duration_2014,chiappa_movement_2010}.

In most of the cases of SDS with an explicit state duration variable, the duration variable is modeled using a categorical variable with support $\{1, 2, \ldots, D_{max}\}$ \citep{ansari_deep_2021,chiappa_explicit-duration_2014,chiappa_movement_2010}.
\section{Problem Statement}
In real-world applications, the data often comes from many independent short trials.
This may happen because the data acquisition process is costly (for instance involves life animals) or the phenomenon we are trying to investigate is bound in time (for instance, sleep).

In real-life scenarios we cannot always control whether the measurements are taken in the way, which allows for unbiased representation of the whole spatiotemporal phenomenon.
For this reason, some regimes in the data can be underrepresented. 

In models without explicit duration variables, self-persistence is strictly related to transition probability, implicitly following the geometric distribution.

On the other hand, numerous natural temporal phenomena display consistent patterns in the duration of a specific model or regime. In these instances, the standard SLDS model fails to accurately capture the regularity of the data. A prime example of this is the honeybee dance, where a dancing bee strives to remain in the waggle regime for a specific period to effectively convey a message. In such cases, it is evident that the actual duration deviates from a geometric distribution.  

rSLDS make transition probability dependent on the position, thus self-persistence no longer follows geometric distribution. However, first-order Markovian recurrence does not adequately address long-term time-dependent switching.

To overcome the problems with rSLDS not capturing the state duration regularity in the data, we introduce recurrent explicit duration variables.
\raggedbottom
\subsection{Model Formulation}
Let's consider the graphical model (presented in the Figure \ref{fig:graphical_model}) defined as
\begin{equation}
\begin{aligned}
&p\left(\mathbf{y}_{1: T}, \mathbf{x}_{1: T}, s_{1: T}, d_{1: T}\right)= p\left(\mathbf{y}_1 \mid \mathbf{x}_1\right)p\left(\mathbf{x}_1 \mid s_1\right)\\
&\quad \cdot p\left(d_1 \mid s_1\right) p\left(s_1\right)\Biggl[\prod_{t=2}^T p\left(\mathbf{y}_t \mid \mathbf{x}_t, s_t\right) p\left(\mathbf{x}_t \mid \mathbf{x}_{t-1}, s_t\right) \\
&\quad \cdot p(s_t \mid s_{t-1}, d_t, \mathbf{x}_{t-1}) p(d_t \mid s_{t}, d_{t-1}, \mathbf{x}_{t-1})\Biggr].
\end{aligned}
\end{equation}

We call this model Recurrent Explicit Duration Switching Linear Dynamical System (REDSLDS).
%The corresponding visual representation is presented in Figure \ref{fig:graphical_model}

Analogously to regular SLDS, the probabilities follow distributions defined below
\begin{equation}%
    \begin{aligned}
p\left(s_1\right) &=\operatorname{Cat}\left(s_1 ; \boldsymbol{\pi_0}\right) \\
p\left(d_1|s_1\right) &=Dur_{s_1, \boldsymbol{\mu}_{s_1}^{init}}(d_1) \\
 p\left(\mathbf{x}_1 \mid s_1\right) &=\mathcal{N}\left(\mathbf{x}_1 ; \boldsymbol{\mu}^{init}_{s_1}, \boldsymbol{\Sigma}^{init}_{s_1}\right) \\
p\left(\mathbf{x}_t \mid \mathbf{x}_{t-1}, s_t\right) &=\mathcal{N}\left(\mathbf{x}_t ; \boldsymbol{A}_{s_{t}}\mathbf{x}_{t-1}+\boldsymbol{a}_{s_{t}}, \boldsymbol{Q}_{s_t}\right) \\
p\left(\mathbf{y}_t \mid \mathbf{x}_{t}, s_t\right) &=\mathcal{N}\left(\mathbf{y}_t ; \boldsymbol{C}_{s_{t}}\mathbf{x}_{t} + \boldsymbol{c}_{s_{t}}, \boldsymbol{S}_{s_t}\right),
\end{aligned}
\end{equation}
where $\pi_0$ is a vector of initial state probabilities, $\operatorname{Cat}$ is categorical pmf, $\operatorname{Dur}$ is duration pmf, $\boldsymbol{\mu}_{s_1}^{init}~:=~ \boldsymbol{A}_{s_{1}}\boldsymbol{\mu}_{s_1}$ and $\boldsymbol{\Sigma}_{s_1}^{init} := \boldsymbol{A}_{s_{1}}\boldsymbol{\Sigma}_{s_1}\boldsymbol{A}_{s_{1}}^{T} + \boldsymbol{Q}_{s_1}$.

The transition of duration $d_t$ and state $s_{t}$ variables is defined as
\begin{equation}
p\left(d_{t} \mid s_{t}, d_{t-1}, \mathbf{x}_{t-1}\right)=\begin{cases}
\delta_{(d_{t-1}-1)} \quad\text { if }  \quad d_{t-1} > 1 \\
Dur_{s_t, \mathbf{x}_{t-1}}(d_t) \quad\text { if }  \quad d_{t-1}=1
\end{cases},
\end{equation}
where $Dur_{s_t, x_{t-1}}$ is duration distribution in state $s_t$, $\delta_{(x)}$ is Dirac's delta, and
\begin{equation}
p\left(s_{t} \mid s_{t-1}, d_{t-1}, \mathbf{x}_{t-1}\right)= \begin{cases}\delta_{s_{t-1}} \quad\text { if }  \quad d_{t-1}>1 \\
\pi_{\mathrm{SB}}(v^{S}_{t}) \quad\text { if }  \quad d_{t-1}=1\end{cases},
\end{equation}
where $\mathbf{v}_{t}^{S}=\mathbf{R}_{s_{t-1}}^{S} \mathbf{x}_{t-1}+\mathbf{r}_{s_{t-1}}$ and $\mathbf{R}_k^S \in \mathbb{R}^{(K-1) \times M}$ is a weight matrix and $\mathbf{r}_k \in \mathbb{R}^{D_{max}-1}$ is a bias vector.

This model formulation allows us to \emph{efficiently represent state durations}.
An alternative way to model explicit durations (employed, for example, by \citet{ansari_deep_2021, oh_learning_2008}) is to use increasing counter variables $c_t$ instead of decreasing duration variables $d_t$. However, we believe that in our case, the duration variables allowed for easier model formulation.

For a comparison of these two modeling approaches, refer to \citet{chiappa_explicit-duration_2014}. 

In our case, 

\begin{equation}
\begin{aligned}
        &Dur_{s_t, \mathbf{x}_{t-1}}= \pi_{\mathrm{SB}}(\mathbf{v}_{t}^{D}), \\
    &\mathbf{v}_{t}^{D}=\mathbf{R}_{s_{t}}^{D} \mathbf{x}_{t-1}+\mathbf{r}_{s_{t}},
\end{aligned}
\end{equation}
where $\mathbf{R}_k^D \in \mathbb{R}^{(D_{max}-1) \times M}$ is a weight matrix and $\mathbf{r}_k \in \mathbb{R}^{D_{max}-1}$ is a bias vector.

We utilized the matrix normal inverse Wishart ($\operatorname{MNIW}(\mathbf{M}, \mathbf{V}, \mathbf{S}, n)$) prior to use for autoregressive dynamics and emissions in each of the experiments. A detailed explanation of the MNIW prior can be found in \citep{fox_bayesian_2011}.
A detailed description of the priors for each of the experiments is provided in the Appendix~\ref{section:priors}.

\subsection{Related Work}

Our model is an extension of the rSLDS model proposed by \citet{linderman_bayesian_2017}.

We note here that it can be regarded as a simpler variant of the REDSDS model proposed by \citet{ansari_deep_2021}. However, their model uses deep neural networks instead of liner models as transition functions.

There are several benefits of our modeling scheme: \emph{we don't have to approximate posterior}, we can use Markov Chain Monte Carlo (MCMC) inference to sample it directly, we do not report problem with ``state-collapse'' as did the models using variational inference \citep{dong_collapsed_2020, ansari_deep_2021}. Our model allows us not only to perform segmentation and prediction, but we are still able to perform smoothing and filtering (see, for example, \citep{fox_bayesian_2011} or algorithms in the Appendix~\ref{section:pseudocodes}).

% We are still able to use Laplace EM or stochastic variational inference, see supplementary material.
\section{Experiments}
\begin{table*}[!h]
\centering
\caption{Results of data segmentation. The means and standard deviations are calculated from ten separate MCMC runs that are independent. We measured \textbf{acc}uracy, \textbf{w}eighted \textbf{F1}, and \textbf{l}og-\textbf{l}ikelihood. The best scores using Init 1 are \underline{underlined}. The best results are \textbf{bold}.}
\label{tab:nascar:results}
\scalebox{.9}{
\begin{tabular}{|l|c|lrrrr|}
\hline

\rowcolor{Gray}
\multicolumn{3}{c}{} & rSLDS (Init I) & REDSLDS (Init I) & rSLDS (Init II) & REDSLDS (Init II) \\
\hline
\multirow{12}{*}{{\rotatebox[origin=c]{90}{\emph{NASCAR$^{\circledR}$} dataset}}} & \multirow{3}{*}{\STAB{\rotatebox[origin=c]{90}{Split 5}}} & Acc      & $0.31 \pm 0.01$ & $\underline{0.63 \pm 0.13}$ & $0.54 \pm 0.08$ & $\mathbf{0.71 \pm 0.01}$ \\
& & Ll & $-1.16  \cdot 10^{12} \pm 5.07  \cdot 10^{11}$ & $\underline{-1.21 \cdot 10^{6} \pm 7.86 \cdot 10^{5}}$ & $7.84 \cdot 10^{4} \pm 1.39 \cdot 10^{4}$ & $\mathbf{8.07 \cdot 10^{4} \pm 2.30 \cdot 10^{3}}$ \\
& & W F1 & $0.33 \pm 0.05$ & $\underline{0.71 \pm 0.08}$ & $0.60 \pm 0.05$ & $\mathbf{0.76 \pm 0.03}$ \\ \cline{2-7}
 &  \multirow{3}{*}{{\rotatebox[origin=c]{90}{\hspace{-1.5mm}Split 10}}}&Acc         & $0.48 \pm 0.02$ & $\mathbf{\underline{0.65 \pm 0.01}}$ & $0.51 \pm 0.01$ & $0.63 \pm 0.07$ \\[1mm]
& & Ll & $\underline{9.13 \cdot 10^{4} \pm 1.01 \cdot 10^{3}}$ & $9.04 \cdot 10^{4} \pm 1.90 \cdot 10^{3}$ & $7.04 \cdot 10^{4} \pm 5.81 \cdot 10^{2}$ & $7.23 \cdot 10^{4} \pm 2.15 \cdot 10^{3}$ \\[1mm]
& & W F1 & $0.49 \pm 0.03$ & $\mathbf{\underline{0.68 \pm 0.02}}$ & $0.49 \pm 0.01$ & $\mathbf{0.68 \pm 0.07}$  \\[1mm] \cline{2-7}
 &  \multirow{3}{*}{\STAB{\rotatebox[origin=c]{90}{Split 15}}}&Acc         & $\underline{0.52 \pm 0.05}$ & $\underline{0.52 \pm 0.09}$ & $0.48 \pm 0.08$ & $\mathbf{0.69 \pm 0.00}$ \\
& & Ll & $\mathbf{\underline{8.20 \cdot 10^{4} \pm 8.85 \cdot 10^{2}}}$ & $7.93 \cdot 10^{4} \pm 1.50 \cdot 10^{3}$ & $6.34 \cdot 10^{4} \pm 1.60  \cdot 10^{3}$ & $6.68  \cdot 10^{4} \pm 7.32 \cdot 10^{2}$ \\
& & W F1 & $\underline{0.55 \pm 0.05}$ & $\underline{0.55 \pm 0.09}$ & $0.54 \pm 0.06$ & $\mathbf{0.73 \pm 0.01}$ \\ \cline{2-7}
 & \multirow{3}{*}{\STAB{\rotatebox[origin=c]{90}{Split 20}}}&Acc         & $0.35 \pm 0.01$ & $\underline{0.42 \pm 0.06}$ & $0.42 \pm 0.06$ & $\mathbf{0.51 \pm 0.05}$ \\
& & Ll & $\mathbf{\underline{4.91 \cdot 10^{4} \pm 6.44 \cdot 10^{3}}}$ & $4.62 \cdot 10^{4} \pm 3.61 \cdot 10^{3}$ &$3.43 \cdot 10^{4} \pm 4.80 \cdot 10^{3}$ & $3.83 \cdot 10^{4} \pm 2.64 \cdot 10^{3}$ \\
& &  W F1 & $0.47 \pm 0.03$ & $0.46 \pm 0.10$ & $0.51 \pm 0.02$ & $\mathbf{0.54 \pm 0.05}$ \\
\hline
\multirow{6}{*}{\STAB{\rotatebox[origin=c]{90}{Bee dataset}}} & \multirow{3}{*}{\STAB{\rotatebox[origin=c]{90}{4,5,6}}} & Acc   & $0.37 \pm 0.01$ & $\underline{\mathbf{0.85 \pm  0.02}}$ & $0.43 \pm 0.03$ & $0.70 \pm 0.03$ \\
& & Ll & $-5.42 \cdot 10^{3} \pm 1.88 \cdot 10^{2}$ & $\underline{-4.73 \cdot 10^{3} \pm 2.99 \cdot 10^{2}}$ & $-3.19 \cdot 10^{8} \pm 3.50 \cdot 10^{8}$ & $\mathbf{-3.03 \cdot 10^{3} \pm 2.79 \cdot 10^{3}}$ \\
& & W F1 & $0.40 \pm 0.00$ & $\underline{\mathbf{0.85 \pm 0.02}}$ & $0.45 \pm 0.04$ & $0.68 \pm 0.02$ \\ \cline{2-7}
& \multirow{3}{*}{\STAB{\rotatebox[origin=c]{90}{All}}} & Acc   & $0.37 \pm 0.02$ & $\underline{0.42 \pm 0.02}$ & $0.40 \pm 0.00$ & $\mathbf{0.62 \pm 0.01}$ \\
 & & Ll &$-1.53  \cdot 10^{2} \pm 2.13 \cdot 10^{10}$ & $\underline{\mathbf{8.62 \cdot 10^{3} \pm 2.62 \cdot 10^{2}}}$ & $-3.80 \cdot 10^{3} \pm 1.27 \cdot 10^{2}$ & $-2.83 \cdot 10^{2} \pm 6.04 \cdot 10^{2}$ \\
& & W F1 & $0.43 \pm 0.05$ & $\underline{0.50 \pm 0.00}$ & $0.57 \pm 0.00$ & $\mathbf{0.62 \pm 0.01}$ \\
\hline
\end{tabular}
}

\end{table*}

\begin{figure*}[!h]
\centering
\hspace*{\fill}%
\subfloat[Split 5 Ground Truth\label{fig:nascar:split5:gt}]{
  \centering
  \includegraphics[width=0.4\linewidth]{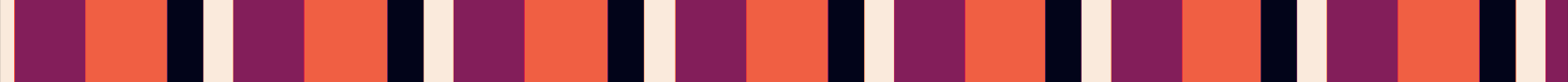}
}\hfill%%
\subfloat[Split 15 Ground Truth\label{fig:nascar:split15:gt}]{
  \centering
  \includegraphics[width=0.4\linewidth]{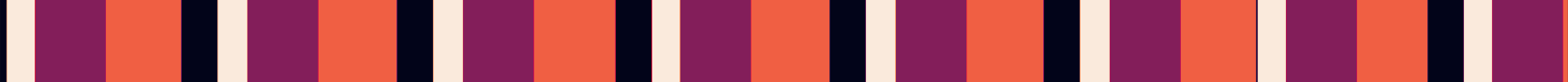}
}\hspace*{\fill}%%%
  \newline
  \hspace*{\fill}%
  \subfloat[Split 5 REDSLDS\label{fig:nascar:split5:redslds}]{
  \centering
  \includegraphics[width=0.4\linewidth]{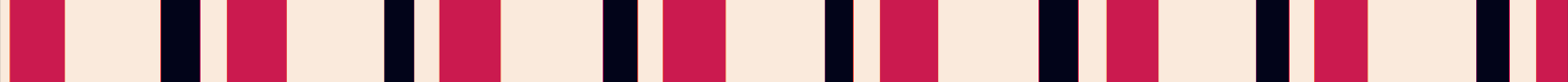}
}%%
\hfill%%
  \subfloat[Split 15 REDSLDS\label{fig:nascar:split15:redslds}]{
  \centering
  \includegraphics[width=0.4\linewidth]{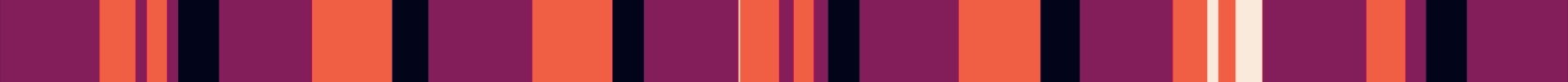}
}%%
\hspace*{\fill}%
\newline
\hspace*{\fill}%
  \subfloat[Split 5 rSLDS\label{fig:nascar:split5:rSLDS}]{
  \centering
  \includegraphics[width=0.4\linewidth]{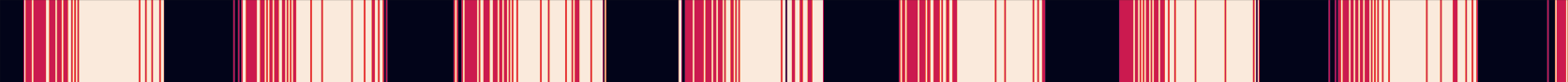}
}%
\hfill%%
\subfloat[Split 15 rSLDS\label{fig:nascar:split15:rSLDS}]{
  \centering
  \includegraphics[width=0.4\linewidth]{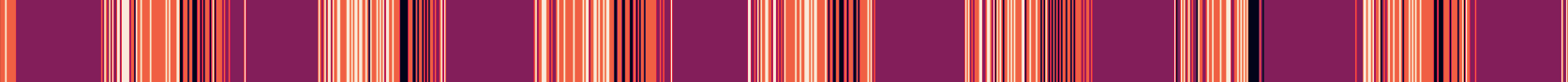}
}%
\hspace*{\fill}%
\newline
\hspace*{\fill}%
\subfloat[Split 10 Ground Truth\label{fig:nascar:split10:gt}]{
  \centering
  \includegraphics[width=0.4\linewidth]{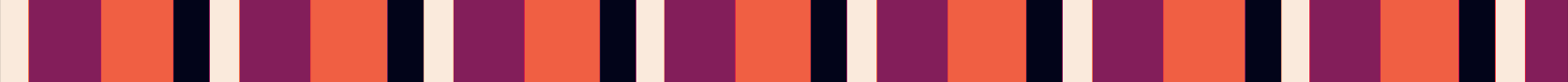}
}%%
\hfill%%
\subfloat[Split 20 Ground Truth\label{fig:nascar:split20:gt}]{
  \centering
  \includegraphics[width=0.4\linewidth]{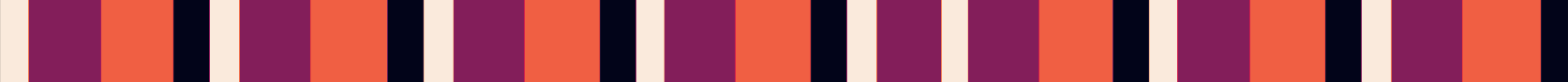}
}%%
\hspace*{\fill}%
  \newline
  \hspace*{\fill}%
  \subfloat[Split 10 REDSLDS\label{fig:nascar:split10:redslds}]{
  \centering
  \includegraphics[width=0.4\linewidth]{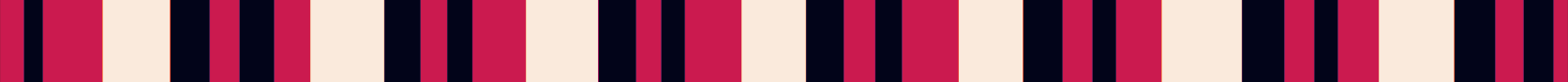}
}%%
\hfill%%
\subfloat[Split 20 REDSLDS\label{fig:nascar:split20:redslds}]{
  \centering
  \includegraphics[width=0.4\linewidth]{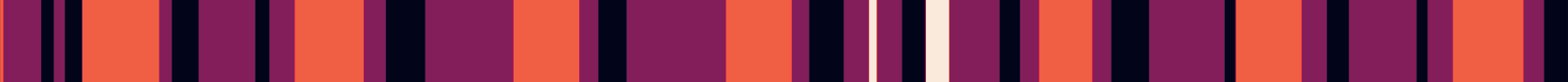}
}%%
\hspace*{\fill}%
\newline
\hspace*{\fill}%
  \subfloat[Split 10 rSLDS\label{fig:nascar:split10:rSLDS}]{
  \centering
  \includegraphics[width=0.4\linewidth]{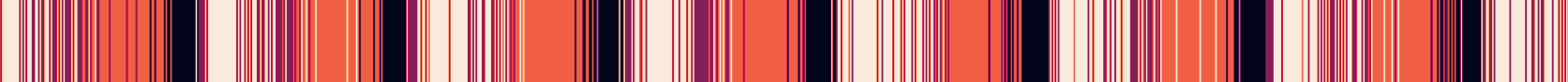}
}%
\hfill%%
\subfloat[Split 20 rSLDS\label{fig:nascar:split20:rSLDS}]{
  \centering
  \includegraphics[width=0.4\linewidth]{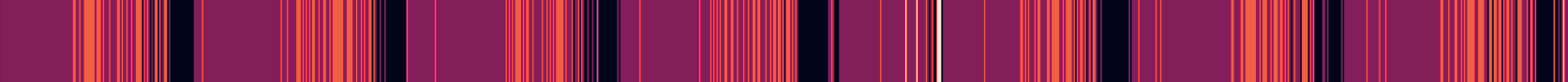}
}%
\hspace*{\fill}%
\caption{NASCAR$^{\circledR}$ segmentations obtained for the $S \in \{5, 10, 15, 20\}$. We can observe that REDSLDS tends to have smoother and more coherent segmentations. However, it tends to ignore the second ``straight-on'' state. }%
\label{fig:nascar:split:segmentation}

\end{figure*}

To assess our model's effectiveness with limited independent observations, we tested it on three benchmarks: a simulated NASCAR$^{\circledR}$ dataset (from \citet{linderman_bayesian_2017}), and two real-world datasets on honeybee dances and mouse behavior. Labels in all experiments were obtained in a fully self-supervised manner.

NASCAR$^{\circledR}$ provides a controlled environment to compare REDSLDS with traditional rSLDS models. The data were sampled to mimic independent samples without clear process periodicity.

The honeybee dataset \citep{oh_learning_2008} features complex dance patterns, making it suitable for testing segmentation in rapid, unstable movements. This aligns with our assumption of independent trials from a shared process.

The mouse behavior dataset \citep{batty_behavenet_2019} presents a high-dimensional challenge, allowing us to test REDSLDS's ability to segment complex behaviors.

These datasets are widely used in the community, including by \citet{linderman_bayesian_2017, nassar_tree-structured_2019, ansari_deep_2021, lee_switching_2023, hutter_disentangled_2021}.

As a baseline, we use rSLDS. Appendix~\ref{section:extendedresults} provides results for SLDS and EDSLDS.

The Python3 implementation of the prototype code utilized the NumPy and SciPy libraries.
% All the calculations were executed on a PC featuring an AMD Ryzen Threadripper 3970X 32-Core Processor.
\paragraph{Initialization} Proper initialization of SLDS-based models is still an open area of research and there are many different schemes \citep{linderman_bayesian_2017, nassar_tree-structured_2019, slupinski_improving_2024}. We initialize the latent continuous state using principal component analysis (PCA), similar to \citet{linderman_bayesian_2017}.
Then we fit five Autoregressive Hidden Markov Models (ARHMMs). Of them, we choose the one with the highest log-likelihood. The states decoded by this model serve as initial state assignment for the REDSLDS model. It was observed that the convergence of the models was quicker when, after their initialization with ARHMM, an additional five iterations were performed with the model parameters kept constant at $z_{1:T}$ and $\mathbf{x}_{1:T}$, a process we denote in Table \ref{tab:nascar:results} as ``Init I''. However, in more complex scenarios, the REDSLDS model showed improvements when the extra iterations were skipped, a phase that we call ``Init II''. This pattern was not significantly evident in simpler models such as rSLDS. This can be attributed to the complex nature of the model, which requires a broader exploration of the parameter space during the initial phases.

The appendices provide detailed priors for each run as well as the sampling scheme.
\begin{figure}[!b]
\centering
\includegraphics[width=0.8\linewidth]{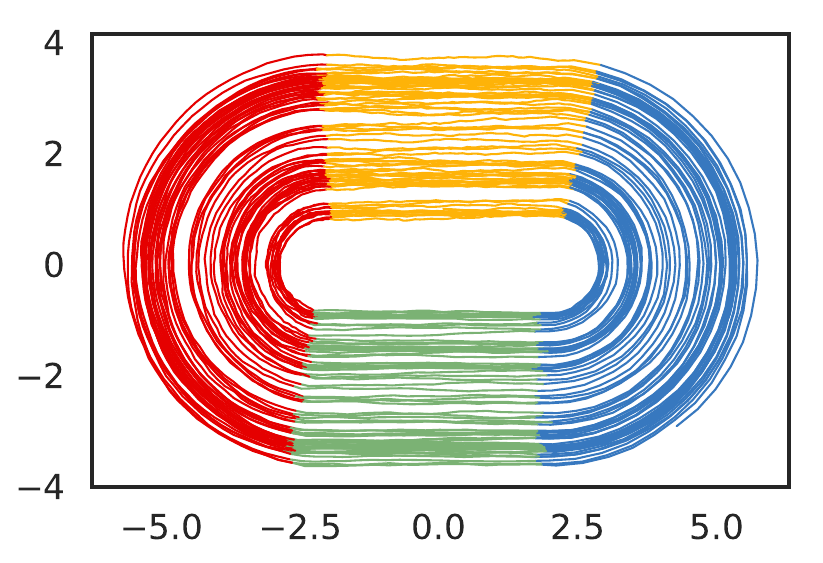}
\caption{Trajectory of \emph{NASCAR$^{\circledR}$} used to test the models.}
\label{fig:nascar:trajectory}
\end{figure}
\subsection{NASCAR$^{\circledR}$}\label{section:nascar}

We begin with a straightforward illustration, where the dynamics exhibit oval shapes resembling those of a stock car on a NASCAR$^{\circledR}$ track\citep{linderman_bayesian_2017} (refer to Figure \ref{fig:nascar:trajectory}). The dynamics is determined by four distinct states, $s_t \in \{1, \ldots, 4\}$, which govern a continuous latent state in two dimensions, $\mathbf{x}_t \in \mathbb{R}^2$. The observations $\mathbf{y}_t \in \mathbb{R}^{10}$, are obtained by linearly projecting the latent state and introducing Gaussian noise.

There is evidence in the literature that rSLDS models perform effectively on this particular kind of data \citep{linderman_bayesian_2017, nassar_tree-structured_2019}. However, we are interested in investigating the performance of SLDS-based models when applied to independent sequences originating from the same distribution. In order to accomplish this, we generated 10 independent runs of NASCAR$^{\circledR}$ with 12000 sample points. We split each run into $S \in {5, 10, 15, 20}$ chunks (multiples of chunks were chosen arbitrarily, there is no reason it should not work for ${3, 6, 12, 24}$ or ${2, 5, 12, 20}$). From the set of the chunks we sampled $N = 0.8S$ chunks, so the whole data set has 9,600 points. Every model was sampled using 10000 Gibbs iterations.

The segmentation quality measures and the log-likelihood scores of the models are shown in Table \ref{tab:nascar:results}. The sample segmentations for rSLDS and REDSLDS are presented in Figure~\ref{fig:nascar:split:segmentation}.

It can be seen that, in general, our model achieved the highest segmentation scores.

We observe that, in general, for each model, the log-likelihood of the model decreases as the number of splits increases (except 5 splits in ``Init I'', but the results suggest that those were underfitted). This enables us to infer that it provides a better fit to the data compared to other models. This aligns with our intuition that shorter sequences pose a greater challenge to all models in accurately representing the data.
The (recurrent)explicit duration variables force the model to stay in a state for a while, whereas in (r)SLDS states can be switched freely.

As can be seen in Table \ref{tab:nascar:results} we benefit both from incorporating recurrence and duration modeling.

\begin{figure}[!htb]
\centering
\hspace*{\fill}%
\subfloat[Ground Truth\label{fig:bee:gt}]{
  \centering
  \includegraphics[width=0.8\linewidth]{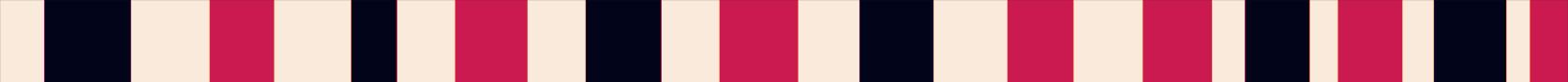}
}%%
\hspace*{\fill}%
  \newline
  \hspace*{\fill}%
  \subfloat[REDSLDS\label{fig:bee:redslds}]{
  \centering
  \includegraphics[width=0.8\linewidth]{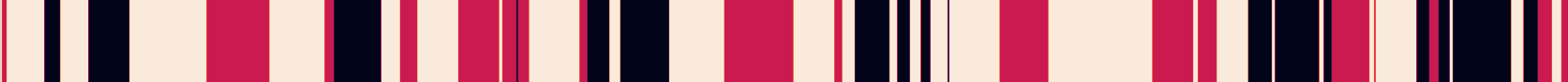}
}%%
\hspace*{\fill}%
\newline
\hspace*{\fill}%
  \subfloat[rSLDS\label{fig:bee:rSLDS}]{
  \centering
  \includegraphics[width=0.8\linewidth]{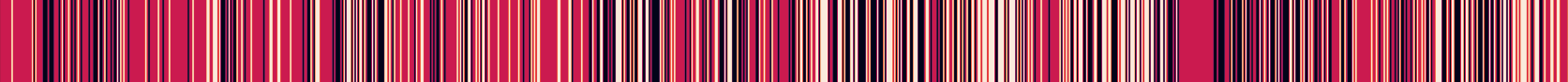}
}%
\hspace*{\fill}%
\caption{Sample segmentations of the dance bee data set. As we can see, rSLDS tends to over-segment the bee movement. It especially struggles with ``waggle'' movement.}%
\label{fig:bee:segmentation}

\end{figure}
\begin{figure*}[h]
\centering
\hspace*{\fill}%
\subfloat[rSLDS\label{fig:behavenet:rslds}]{
  \centering
  \includegraphics[width=0.4\linewidth]{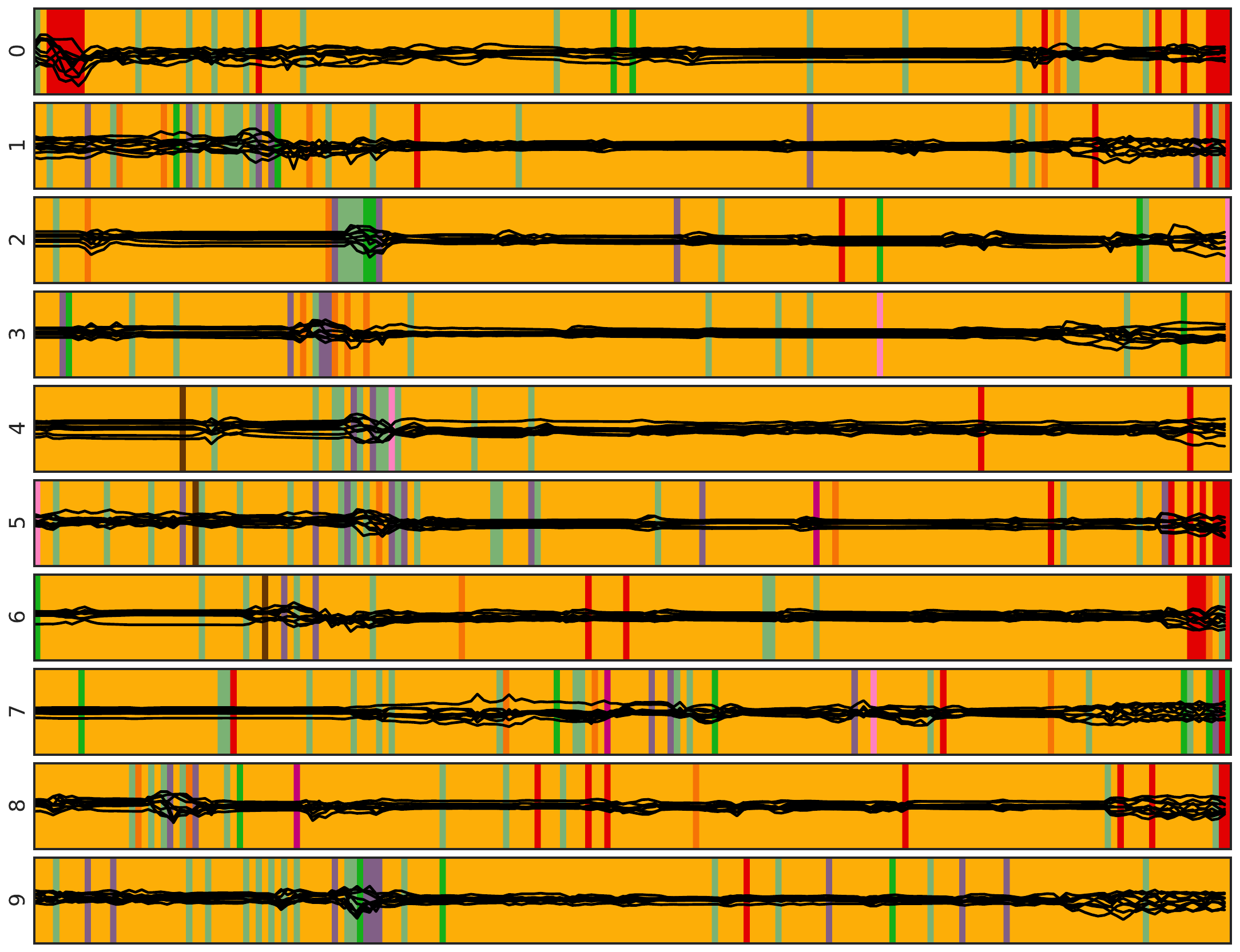}
}%%
\hfill
\subfloat[REDSLDS\label{fig:behavenet:redslds}]{
  \centering
  \includegraphics[width=0.4\linewidth]{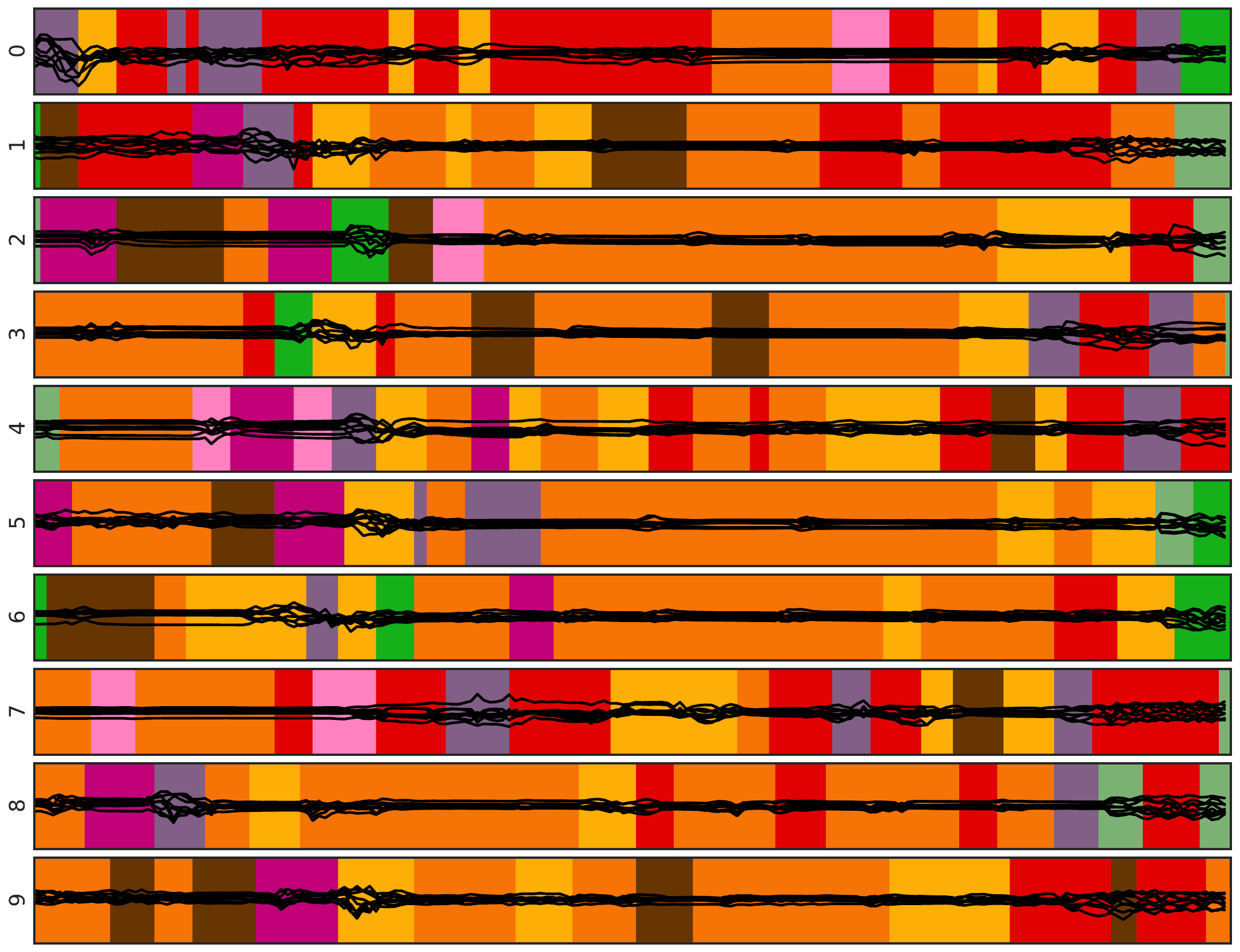}
}%%
\hspace*{\fill}%
\caption{The segmentation of  BehaveNet data obtained by: a) rSLDS model: the figure clearly illustrates that the majority of the data clusters together in a single state, b) REDSLDS model: the presence of a notable degree of self-persistence among the states is apparent. The states exhibit different magnitudes, and the orange state, in particular, demonstrates the least variation in measurement.}
\label{fig:behavenet:segmentation}
\end{figure*}
% \subsection{Lorenz Attractor}
% Lorenz Attractor is one of the most popular benchmarking nonlinear dynamical systems used for benchmarking, appearing in many works regarding state space modeling and switching regime detection \cite{ansari_deep_2021, linderman_bayesian_2017, zoltowski_general_2020, nassar_tree-structured_2019, xu_deep_2021}.
% Its nonlinear dynamics are given by:

% \begin{equation}
% \begin{aligned}
% &\frac{\mathrm{d} x}{\mathrm{~d} t}=\sigma(y-x) \\
% &\frac{\mathrm{d} y}{\mathrm{~d} t}=x(\rho-z)-y \\
% &\frac{\mathrm{d} z}{\mathrm{~d} t}=x y-\beta z .
% \end{aligned}
% \end{equation}

% We trained models on 5000 timepoints sampled for $\sigma = 10$, $\rho = 28$, $\beta=2.667$.
% For the reasons outlined above, we decided to use three hidden states.

\subsection{Dancing Bee}\label{section:bee}
We utilized the publicly accessible dancing bee dataset \citep{oh_learning_2008}, which is known for its complex nature and has previously been examined in the realm of time series segmentation. 

The data set contains information on the movements of six honeybees as they perform the waggle dance. This includes their coordinates in a 2D space and the angle at which they are heading at each time interval.

Communication of food sources among honey bees is accomplished through a series of dances performed within the hive. These dances involve specific movements, such as waggle, turn right, and turn left. The waggle dance involves the bee walking in a straight line while rapidly shaking its body from side to side. The turning dances simply involve the bee rotating clockwise or counterclockwise. The data collected for analysis consists of $\boldsymbol{y}_t=\left[\begin{array}{llll}\cos \left(\theta_t\right), & \sin \left(\theta_t\right), & x_t, & y_t\end{array}\right]^T$, where $\left(x_t, y_t\right)$ represents the two-dimensional coordinates of the body of the bee and $\theta_t$ corresponds to the angle of its head.

Since the data set comprises separate recordings of bee dances, it aligns with our assumption that the data consists of independent trials from a shared process. 

We scored models on both the whole dataset and the subset of the last three subsequences, since the first three have slightly different characteristics and lengths. Every model was sampled using 10000 Gibbs iterations.

The Table \ref{tab:nascar:results} displays the complete segmentation scores and model log-likelihoods. It can be seen that our model outperforms all of the other baselines. For comparison, REDSDS for \citet{ansari_deep_2021} scored the precision of ($0.73\pm0.10$). Figure \ref{fig:bee:segmentation} presents the sample segmentations.

The rSLDS algorithm faces challenges in capturing long-term motion patterns, resulting in a large number of segments. This problem is especially noticeable during the ``waggle'' phase of the dance, which is characterized by rapid and unsteady movements, as shown in Figure \ref{fig:bee:rSLDS}.
\subsection{BehaveNet}\label{section:behavenet}
We also used our model on a publicly available dataset of mouse behavior \citep{batty_behavenet_2019}. In this particular experiment, a mouse with a fixed head position performed a visual decision task while neural activity in the dorsal cortex was recorded using wide-field calcium imaging. We focused solely on the behavior video data, which consisted of gray-scale video frames with dimensions of 128x128 pixels. The behavior video was captured using two cameras, one providing a side view and the other providing a bottom view. Due to the high dimensionality of the behavior video, we directly utilized the dimension reduction results from a previous study. These results consisted of 9-dimensional continuous variables estimated using a convolutional autoencoder.
\emph{As there are no labels in this dataset, we treat it as a qualitative experiment.} Every model was sampled using 5000 Gibbs iterations.

For this experiment, we only used rSLDS and REDSLDS, using initialization scheme 2.
Of the whole dataset, we sample ten sequences with uniform probability. We standardize the measurements.

The use of ARHMM for segmentation in this data set is susceptible to over-segmentation, a concern that has recently been tackled by \citep{costacurta_distinguishing_2022}. In their study, the authors introduced auxiliary variables that added additional noise to the conventional ARHMM model.

In models based on SLDS, we introduce extra noise to the autoregression process through the emission layer.

The segmentation obtained by the rSLDS model is shown in Figure \ref{fig:behavenet:rslds}. This model achieved the highest log-likelihood of $-13420.80$. It can be seen from the figure that most of the data converges to a single state.

The segmentation achieved by the REDSLDS model is shown in Figure \ref{fig:behavenet:redslds}, with a log-likelihood value of $-13245.71$. It can be seen that there is a significant level of self-persistence among states. The various states clearly exhibit distinct amplitudes, with the orange state particularly displaying the lowest change in measurement.

\section{Conclusions}
In this paper, we proposed REDSLDS, being an extension of rSLDS with recurrent explicit duration variables, as well as the inference and learning scheme with Pólya-gamma augmentation. Adding time-dependent switching supported state-dependent switching from the original rSLDS and led to a significant improvement of the learning process. As computational experiments showed, our model significantly outperformed the segmentation capabilities of the original rSLDS on three benchmark datasets, usually used in the evaluation of the SLDS models.

% In this paper we introduced REDSLDS - a new model introducing recurrent explicit duration variables to rSLDS model. We introduced inference and learning scheme involving introduction of Pólya-gamma augmentation. We have shown how our model enhances segmentation capabiliteis of rSLDS on three benchmark datasets.

The primary novelty presented in this paper is the incorporation of recurrent explicit duration random variables in (r)SLDS. The utility of these variables was previously demonstrated by \citet{ansari_deep_2021}, who extended SNLDS by \citet{dong_collapsed_2020} in a manner similar to our extension of REDSLDS.

In the Appendix~\ref{section:formulas}, we share our computations and details of the message passing scheme, both for REDSLDS and rSLDS, to facilitate easy reproduction of our work.

We also examined the impact of the length of the sampled spatiotemporal data on the learning process in switching dynamical models, which we believe has not been done before.

We evaluated our model using three well-established benchmarks in the community and demonstrated its effectiveness.

Future work may involve evaluation on more complex datasets.
Since our model has shown an improvement in performance over rSLDS in the segmentation task, it is worth investigating applications of this model in tasks where rSLDS or SLDS are applied. Those may involve neural data modeling \citep{zoltowski_general_2020, osuna-orozco_identification_2023, song_unsupervised_2023, jiang_dynamic_2024, bush_latent_2024}, early warning of systemic banking crises \citep{dabrowski_systemic_2016}, or monitoring of neonatal conditions~\citep{stanculescu_hierarchical_2014}.

\section*{Acknowledgements}
Calculations have been carried out using resources provided by Wroclaw Centre for Networking and Supercomputing (http://wcss.pl), grant No. 405. 

This work was supported by the Polish National Science Centre (NCN) under grant OPUS-18 no. 2019/35/B/ST6/04379.
% In this study, we only considered categorical distribution for our explicit duration variable, but choosing one with infinite support like \citet{johnson_bayesian_2013}, creating semiparametric model would also be an interesting idea.

% Another interesting direction is the nonparametric version of this model, based on HDP-SLDS \citep{fox_bayesian_2011}.

%%% Use this command to include your bibliography file.
\newpage
\bibliography{bibliography}
\onecolumn
\appendix

\section{Conditional posteriors}

The structure of the model allows for closed-form conditional posterior distributions that are easy to sample from. For clarity, the conditional posterior distributions for the REDSLDS are given below:

The linear dynamic parameters $\left(A_k, a_k\right)$ and state variance $Q_k$ of a mode $k$ are conjugate with a Matrix Normal Inverse Wishart (MNIW) prior
\begin{equation}
p\left(\left(\boldsymbol{A}_k, \boldsymbol{a}_k\right), \boldsymbol{Q}_k \mid \boldsymbol{x}_{0: T}, s_{1: T}\right) \propto p\left(\left(\boldsymbol{A}_k, \boldsymbol{b}_k\right), \boldsymbol{Q}_k\right) \prod_{t=1}^T \mathcal{N}\left(\boldsymbol{x}_t \mid \boldsymbol{x}_{t-1}+\boldsymbol{A}_{s_t} \boldsymbol{x}_{t-1}+\boldsymbol{a}_{s_t}, \boldsymbol{Q}_{s_t}\right)^{\mathbb{I}\left[s_t=k\right]} .
\end{equation}

If we assume the observation model is linear and with additive white Gaussian noise then the emission parameters $\Psi=\{(\boldsymbol{C}_{k}, \boldsymbol{c}_{k}), \boldsymbol{S}_k\}$ are also conjugate with a MNIW prior
\begin{equation}
p\left((\boldsymbol{C}_{s_t}, \boldsymbol{c}_{s_t}), \boldsymbol{S}_{s_t} \mid \boldsymbol{x}_{1: T}, \boldsymbol{y}_{1: T}, s_{1: T}\right) \propto p((\boldsymbol{C}_{s_t}, \boldsymbol{c}_{s_t}), \boldsymbol{S}_{s_t}) \prod_{t=1}^T \mathcal{N}\left(\boldsymbol{y}_t \mid \boldsymbol{C}_{s_t} \boldsymbol{x}_t+\boldsymbol{c}_{s_t}, \boldsymbol{S}_{s_t}\right)^{\mathbb{I}\left[s_t=k\right]} .
\end{equation}

We can express the posterior over the hyperplane of a state as:
\begin{equation}
p\left(\left(\boldsymbol{R}_k^S, \boldsymbol{r}_k^S\right) \mid \boldsymbol{x}_{0: T}, s_{1: T}, \omega^S_{k, 1: T}\right) \propto p\left(\left(\boldsymbol{R}_k^S, \boldsymbol{r}_k^S\right)\right) \prod_{t=1}^T \mathcal{N}\left(\boldsymbol{v}_{k, t}^S \mid \kappa_{k, t}^S / \omega_{k, t}^S, 1 / \omega_{k, t}^S\right)^{\mathbb{I}\left[s_t=k\right]},
\end{equation}
where $\kappa_{k, t}=\mathbb{I}[s_t=k]-\frac{1}{2} \mathbb{I}[s_t \geq k]$. Augmenting the model with Pólya-gamma random variables allows for the posterior to be conditionally Gaussian under a Gaussian prior.

Analogously,
\begin{equation}
p\left(\left(\boldsymbol{R}_k^D, \boldsymbol{r}_k^D\right) \mid \boldsymbol{x}_{0: T}, d_{1: T}, s_{1: T}, \omega^D_{d, 1: T}\right) \propto p\left(\left(\boldsymbol{R}_k^D, \boldsymbol{r}_k^D\right)\right) \prod_{t=1}^T \mathcal{N}\left(\boldsymbol{v}_{k, t}^D \mid \kappa_{d, t}^D / \omega_{d, t}^D, 1 / \omega_{d, t}^D\right)^{\mathbb{I}\left[s_t=k\right]},
\end{equation}
where $\kappa_{d, t}=\mathbb{I}[d_t=d]-\frac{1}{2} \mathbb{I}[d_t \geq d]$.

The conditional posterior of the Pólya-gamma random variables are also Pólya-gamma:
\begin{equation}\label{eq:pq_posterior}
    \omega_{n, t} \mid s_t,\left(\boldsymbol{R}_n, \boldsymbol{r}_n\right), \boldsymbol{x}_{t-1} \sim \operatorname{PG}\left(1, \boldsymbol{v}_{n, t}\right).
\end{equation}

\section{Message passing for recurrent connections}\label{section:formulas}

To still use Kalman filtering, we have to compute backward messages incorporating information from sampled durations and states. Assume $m_{t+1, t}\left(\boldsymbol{x}_{t}\right) \propto \mathcal{N}^{-1}\left(\boldsymbol{x}_{t} ; \boldsymbol{\theta}_{t+1, t}, \boldsymbol{\Lambda}_{t+1, t}\right)$, where $\mathcal{N}^{-1}(\boldsymbol{x} ; \boldsymbol{\theta}, \boldsymbol{\Lambda})$ denotes a Gaussian distribution on $x$ in information form with mean $\boldsymbol{\mu}=\boldsymbol{\Lambda}^{-1} \boldsymbol{\theta}$ and covariance $\boldsymbol{\Sigma}=\boldsymbol{\Lambda}^{-1}$. Given a fixed mode sequence $s_{1: T}$ and duration sequence $s_{1: T}$ we simply have a time-varying linear dynamic system. The backward messages can be recursively defined by
\begin{equation}
m_{t, t-1}\left(\boldsymbol{x}_{t-1}\right) \propto \int_{\mathcal{X}_{t}} p\left(\boldsymbol{x}_{t} \mid \boldsymbol{x}_{t-1}, s_{t}\right) p\left(\boldsymbol{y}_{t} \mid \boldsymbol{x}_{t}, s_t\right) p\left(d_{t} \mid \boldsymbol{x}_{t-1}, s_{t}, d_{t-1}\right) p\left(s_{t} \mid \boldsymbol{x}_{t}, s_{t-1}, d_{t-1}\right) m_{t+1, t}\left(\boldsymbol{x}_{t}\right) d \boldsymbol{x}_{t}
\end{equation}

The transition density may be expressed as

\begin{equation}
\begin{aligned}
&p\left(\boldsymbol{x}_{t} \mid \boldsymbol{x}_{t-1}, s_{t}\right) \propto \exp \left\{-\frac{1}{2}\left(\boldsymbol{x}_{t}-\mathbf{A}_{s_{t}} \boldsymbol{x}_{t-1} - \mathbf{a}_{s_t}\right)^{T} \mathbf{Q}^{-1}_{s_{t}}\left(\boldsymbol{x}_{t}-\mathbf{A}_{s_{t}} \boldsymbol{x}_{t-1} - \mathbf{a}_{s_t}\right)\right\}\\
&\propto \exp \left\{-\frac{1}{2}\left[\begin{array}{c}
\boldsymbol{x}_{t-1} \\
\boldsymbol{x}_{t}
\end{array}\right]^{T}\left[\begin{array}{cc}
\mathbf{A}_{s_{t}}^{T} \mathbf{Q}^{-1}_{s_{t}} \mathbf{A}_{s_{t}} & -\mathbf{A}_{s_{t}}^{T} \mathbf{Q}^{-1}_{s_{t}} \\
-\mathbf{Q}^{-1}_{s_{t}} \mathbf{A}_{s_{t}} & \mathbf{Q}^{-1}_{s_{t}}
\end{array}\right]\left[\begin{array}{c}
\boldsymbol{x}_{t-1} \\
\boldsymbol{x}_{t}
\end{array}\right] +\left[\begin{array}{c}
\boldsymbol{x}_{t-1} \\
\boldsymbol{x}_{t}
\end{array}\right]^{T}\left[\begin{array}{cc}
-\mathbf{A}_{s_{t}}^{T} \mathbf{Q}^{-1}_{s_{t}} \mathbf{a}_{s_t} \\
\mathbf{Q}^{-1}_{s_{t}} \mathbf{a}_{s_t}
\end{array}\right]\right\},
\end{aligned}
\end{equation}

The likelihood is given by:
\begin{equation}
\begin{aligned}
p\left(\boldsymbol{y}_{t} \mid \boldsymbol{x}_{t}, s_t\right) & \propto \exp \left\{-\frac{1}{2}\left(\boldsymbol{y}_{t}-\mathbf{C}_{s_t} \boldsymbol{x}_{t} - \boldsymbol{c}_{s_t}\right)^{T} \mathbf{S}_{s_{t}}^{-1}\left(\boldsymbol{y}_{t}-\mathbf{C}_{s_t} \boldsymbol{x}_{t} - \boldsymbol{c}_{s_t}\right)\right\} \\
& \propto \exp \left\{-\frac{1}{2}\left[\begin{array}{c}
\boldsymbol{x}_{t-1} \\
\boldsymbol{x}_{t}
\end{array}\right]^{T}\left[\begin{array}{cc}
\boldsymbol{0} & \boldsymbol{0} \\
\boldsymbol{0} & \mathbf{C}_{s_t}^{T} \mathbf{S}_{s_{t}}^{-1} \mathbf{C}_{s_t}
\end{array}\right]\left[\begin{array}{c}
\boldsymbol{x}_{t-1} \\
\boldsymbol{x}_{t}
\end{array}\right]+\left[\begin{array}{c}
\boldsymbol{x}_{t-1} \\
\boldsymbol{x}_{t}
\end{array}\right]^{T}\left[\begin{array}{cc}
\boldsymbol{0} \\
\mathbf{C}_{s_t}^{T} \mathbf{S}_{s_{t}}^{-1} \boldsymbol{y}_{t} - \mathbf{C}_{s_t}^{T} \mathbf{S}_{s_{t}}^{-1}\boldsymbol{c}_{s_t}
\end{array}\right]\right\},
\end{aligned}
\end{equation}
Due to the addition of recurrent connections, we have to condition $\boldsymbol{x_{t-1}}$ on the future discrete state $s_{t}$:
\begin{equation}
\begin{aligned}
p\left(s_{t} \mid \boldsymbol{x}_{t-1}, s_{t-1}, d_{t-1}\right) & \propto \exp \left\{-\frac{\mathbb{I}[d_{t-1} = 1]}{2}\left(\boldsymbol{v}^{S}_{t}-\boldsymbol{\Omega}^{-S}_{t-1} \boldsymbol{\kappa}_{t}^S\right)^{T} \boldsymbol{\Omega}^{S}_{t-1}\left(\boldsymbol{v}^{S}_{t}-\boldsymbol{\Omega}^{-S}_{t-1} \boldsymbol{\kappa}_{t}^S\right)\right\} \\
& \propto \exp \left\{-\frac{\mathbb{I}[d_{t-1} = 1]}{2}\left[\begin{array}{c}
\boldsymbol{x}_{t-1} \\
\boldsymbol{x}_{t}
\end{array}\right]^{T}\left[\begin{array}{cc}
\boldsymbol{\Omega}^{S}_{t-1} \mathbf{R}_{s_{t-1}}^{S T}\mathbf{R}_{s_{t-1}}^S & \boldsymbol{0} \\
\boldsymbol{0} & \boldsymbol{0}
\end{array}\right]\left[\begin{array}{c}
\boldsymbol{x}_{t-1} \\
\boldsymbol{x}_{t}
\end{array}\right] \right\} \\
&\cdot \exp\left\{\mathbb{I}[d_{t-1} = 1]\left[\begin{array}{c}
\boldsymbol{x}_{t-1} \\
\boldsymbol{x}_{t}
\end{array}\right]^{T}\left[\begin{array}{cc}
\mathbf{R}_{s_{t-1}}^{S T}\boldsymbol{\kappa}^S_{t} - \mathbf{R}_{s_{t-1}}^S \boldsymbol{\Omega}^{S}_{t-1} \boldsymbol{r}_{s_{t-1}}^S \\
\boldsymbol{0}
\end{array}\right]\right\},
\end{aligned}
\end{equation}
with $\boldsymbol{\Omega}^{S}_{t-1}=\operatorname{diag}\left(\omega_{t-1}\right)$ and $\kappa_{t}^S=\left[\kappa_{t,1}^S \ldots, \kappa_{t, K-1}^S\right]$.

Also conditioning $\boldsymbol{x_{t-1}}$ on the future duration time $d_{t}$ is required:
\begin{equation}
\begin{aligned}
p\left(d_t \mid \boldsymbol{x}_{t-1}, s_{t}, d_{t-1}\right) & \propto \exp \left\{-\frac{\mathbb{I}[d_{t-1} = 1]}{2}\left(\boldsymbol{v}^{D}_{t}-\boldsymbol{\Omega}^{-D}_{t-1} \boldsymbol{\kappa}_{t}^D\right)^{T} \boldsymbol{\Omega}^{D}_{t-1}\left(\boldsymbol{v}^{D}_{t}-\boldsymbol{\Omega}^{-D}_{t-1} \boldsymbol{\kappa}_{t}^D\right)\right\} \\
& \propto \exp \left\{-\frac{\mathbb{I}[d_{t-1} = 1]}{2}\left[\begin{array}{c}
\boldsymbol{x}_{t-1} \\
\boldsymbol{x}_{t}
\end{array}\right]^{T}\left[\begin{array}{cc}
\boldsymbol{\Omega}^{D}_{t-1} \mathbf{R}_{s_{t}}^{D T}\mathbf{R}_{s_{t}}^D & \boldsymbol{0} \\
\boldsymbol{0} & \boldsymbol{0}
\end{array}\right]\left[\begin{array}{c}
\boldsymbol{x}_{t-1} \\
\boldsymbol{x}_{t}
\end{array}\right] \right\} \\ 
&\cdot \exp\left\{\mathbb{I}[d_{t-1} = 1]\left[\begin{array}{c}
\boldsymbol{x}_{t-1} \\
\boldsymbol{x}_{t}
\end{array}\right]^{T}\left[\begin{array}{cc}
\mathbf{R}_{s_{t}}^{D T}\boldsymbol{\kappa}^D_{t} - \mathbf{R}_{s_{t}}^D \boldsymbol{\Omega}^{D}_{t-1} \boldsymbol{r}_{s_{t}}^D \\
\boldsymbol{0}
\end{array}\right]\right\},
\end{aligned}
\end{equation}
with $\boldsymbol{\Omega}^{D}_{t-1}=\operatorname{diag}\left(\omega_{t-1}\right)$ and $\boldsymbol{\kappa_{t}}^D=\left[\kappa_{t,1}^D \ldots, \kappa_{t, D_{max}-1}^D\right]$.

The product of these quadratics is given by:

\begin{equation}
\begin{aligned}
&p\left(\boldsymbol{x}_{t} \mid \boldsymbol{x}_{t-1}, s_{t}\right) p\left(\boldsymbol{y}_{t} \mid \boldsymbol{x}_{t}, s_t\right) p\left(d_{t} \mid \boldsymbol{x}_{t-1}, s_{t}, d_{t-1}\right) p\left(s_{t} \mid \boldsymbol{x}_{t}, s_{t-1}, d_{t-1}\right) m_{t+1, t}\left(\boldsymbol{x}_{t}\right) \propto\\
&\exp \left\{-\frac{1}{2}\left[\begin{array}{c}
\boldsymbol{x}_{t-1} \\
\boldsymbol{x}_{t}
\end{array}\right]^{T}\left[\begin{array}{cc}
\mathbf{A}_{s_t}^{T} \mathbf{Q}_{s_t}^{-1} \mathbf{A} + \mathbb{I}[d_{t-1} = 1](\boldsymbol{\Omega}^{S}_{t-1} \mathbf{R}_{s_{t-1}}^{S T}\mathbf{R}_{s_{t-1}}^S + \boldsymbol{\Omega}^{D}_{t-1} \mathbf{R}_{s_{t}}^{D T}\mathbf{R}_{s_{t}}^D) & -\mathbf{A}_{s_t}^{T} \mathbf{Q}_{s_t}^{-1} \\
-\mathbf{Q}_{s_t}^{-1} \mathbf{A}_{s_{t}} & \mathbf{Q}_{s_t}^{-1}+\mathbf{C}_{s_t}^{T} \mathbf{S}_{s_{t}}^{-1} \mathbf{C}_{s_t}+\boldsymbol{\Lambda}_{t+1, t}
\end{array}\right]\left[\begin{array}{c}
\boldsymbol{x}_{t-1} \\

\boldsymbol{x}_{t}
\end{array}\right]\right\} \\
&\cdot \exp\left\{\left[\begin{array}{c}
\boldsymbol{x}_{t-1} \\
\boldsymbol{x}_{t}
\end{array}\right]^{T}\left[\begin{array}{cc}
\mathbb{I}[d_{t-1} = 1](\mathbf{R}_{s_{t-1}}^{D T}\boldsymbol{\kappa}^D_{t} - \mathbf{R}_{s_{t-1}}^D \boldsymbol{\Omega}^{D}_{t-1} \boldsymbol{r}_{s_{t}}^D + \mathbf{R}_{s_{t}}^{S T}\boldsymbol{\kappa}^S_{t} - \mathbf{R}_{s_{t-1}}^S \boldsymbol{\Omega}^{S}_{t-1} \boldsymbol{r}_{s_{t-1}}^S) -\mathbf{A}_{s_t}^{T} \mathbf{Q}_{s_t}^{-1} \mathbf{a}_{s_t}\\
 \boldsymbol{\theta}_{t+1, t} + \mathbf{C}_{s_t}^{T} \mathbf{S}_{s_{t}}^{-1} \boldsymbol{y}_{t} - \mathbf{C}_{s_t}^{T} \mathbf{S}_{s_{t}}^{-1}\boldsymbol{c}_{s_t} +\mathbf{Q}_{s_t}^{-1} \mathbf{a}_{s_t}

\end{array}\right]
\right\},
\end{aligned}
\end{equation}

Now we want to marginalize everything to obtain:
\begin{equation}
m_{t, t-1}\left(\boldsymbol{x}_{t-1}\right) \propto \mathcal{N}^{-1}\left(\boldsymbol{x}_{t-1} ; \boldsymbol{\theta}_{t, t-1}, \boldsymbol{\Lambda}_{t, t-1}\right)
\end{equation}

which is possible using the identity:

\begin{equation}
\int_{\mathcal{X}_{2}} \mathcal{N}^{-1}\left(\left[\begin{array}{l}
\boldsymbol{x}_{1} \\
\boldsymbol{x}_{2}
\end{array}\right] ;\left[\begin{array}{l}
\boldsymbol{\theta}_{1} \\
\boldsymbol{\theta}_{2}
\end{array}\right],\left[\begin{array}{cc}
\boldsymbol{\Lambda}_{11} & \boldsymbol{\Lambda}_{12} \\
\boldsymbol{\Lambda}_{21} & \boldsymbol{\Lambda}_{22}
\end{array}\right]\right) d \boldsymbol{x}_{2}=\mathcal{N}^{-1}\left(x_{1} ; \boldsymbol{\theta}_{1}-\boldsymbol{\Lambda}_{12} \boldsymbol{\Lambda}_{22}^{-1} \boldsymbol{\theta}_{2}, \boldsymbol{\Lambda}_{11}-\boldsymbol{\Lambda}_{12} \boldsymbol{\Lambda}_{22}^{-1} \boldsymbol{\Lambda}_{21}\right)
\end{equation}

Finally we get:
\begin{align}
\boldsymbol{\theta}_{t, t-1}&= \mathbb{I}[d_{t-1} = 1](\mathbf{R}_{s_{t}}^{D T}\boldsymbol{\kappa}^D_{t} - \mathbf{R}_{s_{t}}^D \boldsymbol{\Omega}^{D}_{t-1} \boldsymbol{r}_{s_{t}}^D + \mathbf{R}_{s_{t-1}}^{S T}\boldsymbol{\kappa}^S_{t} - \mathbf{R}_{s_{t-1}}^S \boldsymbol{\Omega}^{S}_{t-1} \boldsymbol{r}_{s_{t-1}}^S) -\mathbf{A}_{s_t}^{T} \mathbf{Q}_{s_t}^{-1} \mathbf{a}_{s_t} \nonumber \\
&+\mathbf{A}_{s_t}^{T} \mathbf{Q}_{s_t}^{-1}(\mathbf{Q}_{s_t}^{-1}+\mathbf{C}_{s_t}^{T} \mathbf{S}_{s_{t}}^{-1} \mathbf{C}_{s_t}+\boldsymbol{\Lambda}_{t+1, t})^{-1}(\boldsymbol{\theta}_{t+1, t} + \mathbf{C}_{s_t}^{T} \mathbf{S}_{s_{t}}^{-1} \boldsymbol{y}_{t} - \mathbf{C}_{s_t}^{T} \mathbf{S}_{s_{t}}^{-1}\boldsymbol{c}_{s_t} +\mathbf{Q}_{s_t}^{-1} \mathbf{a}_{s_t})
\\
\boldsymbol{\Lambda}_{t, t-1}&= \mathbf{A}_{s_t}^{T} \mathbf{Q}_{s_t}^{-1} \mathbf{A} + \mathbb{I}[d_{t-1} = 1](\boldsymbol{\Omega}^{S}_{t-1} \mathbf{R}_{s_{t-1}}^{S T}\mathbf{R}_{s_{t-1}}^S + \boldsymbol{\Omega}^{D}_{t} \mathbf{R}_{s_{t}}^{D T}\mathbf{R}_{s_{t}}^D) \nonumber \\
&-\mathbf{A}_{s_t}^{T}\mathbf{Q}_{s_t}^{-1} (\mathbf{Q}_{s_t}^{-1}+\mathbf{C}_{s_t}^{T} \mathbf{S}_{s_{t}}^{-1} \mathbf{C}_{s_t}+\boldsymbol{\Lambda}_{t+1, t})^{-1}\mathbf{Q}_{s_t}^{-1} \mathbf{A}_{s_{t}}
\end{align}
We initialize the recursion with:
\begin{equation}
m_{T+1, T} \sim \mathcal{N}^{-1}\left(\boldsymbol{x}_{T} ; \boldsymbol{0},\boldsymbol{0}\right)
\end{equation}

Let,
\begin{equation}
\begin{aligned}
\boldsymbol{\Lambda}_{t \mid t}^{b} &=\mathbf{C}_{s_t}^{T} \mathbf{S}_{s_{t}}^{-1} \mathbf{C}_{s_t}+\boldsymbol{\Lambda}_{t+1, t} \\
\boldsymbol{\theta}_{t \mid t}^{b} &=\mathbf{C}_{s_t}^{T} \mathbf{S}_{s_{t}}^{-1} \boldsymbol{y}_{t} - \mathbf{C}_{s_t}^{T} \mathbf{S}_{s_{t}}^{-1}\boldsymbol{c}_{s_t}+\boldsymbol{\theta}_{t+1, t}
\end{aligned}
\end{equation}
Then we can define the following recursion, which we note is equivalent to a backwards running Kalman,
\begin{equation}
\begin{aligned}
\boldsymbol{\Lambda}_{t-1 \mid t-1}^{b}=& \mathbf{C}_{s_{t-1}}^{T} \mathbf{S}_{s_{t-1}}^{-1} \mathbf{C}_{s_{t-1}}+\mathbf{A}_{s_t}^{T} \mathbf{Q}_{s_t}^{-1} \mathbf{A} + \mathbb{I}[d_{t-1} = 1](\boldsymbol{\Omega}^{S}_{t-1} \mathbf{R}_{s_{t-1}}^{S T}\mathbf{R}_{s_{t-1}}^S + \boldsymbol{\Omega}^{D}_{t-1} \mathbf{R}_{s_{t}}^{D T}\mathbf{R}_{s_{t}}^D) \nonumber \\
&-\mathbf{A}_{s_t}^{T}\mathbf{Q}_{s_t}^{-1} (\mathbf{Q}_{s_t}^{-1}+\mathbf{C}_{s_t}^{T} \mathbf{S}_{s_{t}}^{-1} \mathbf{C}_{s_t}+\boldsymbol{\Lambda}_{t+1, t})^{-1}\mathbf{Q}_{s_t}^{-1} \mathbf{A}_{s_{t}} \\
=& \mathbf{C}_{s_{t-1}}^{T} \mathbf{S}_{s_{t-1}}^{-1} \mathbf{C}_{s_{t-1}}+ \mathbf{A}_{s_t}^{T} \mathbf{Q}_{s_t}^{-1} \mathbf{A} + \mathbb{I}[d_{t-1} = 1](\boldsymbol{\Omega}^{S}_{t-1} \mathbf{R}_{s_{t-1}}^{S T}\mathbf{R}_{s_{t-1}}^S + \boldsymbol{\Omega}^{D}_{t-1} \mathbf{R}_{s_{t}}^{D T}\mathbf{R}_{s_{t}}^D) \nonumber \\
&-\mathbf{A}_{s_t}^{T}\mathbf{Q}_{s_t}^{-1} (\mathbf{Q}_{s_t}^{-1}+\boldsymbol{\Lambda}_{t\mid t})^{-1}\mathbf{Q}_{s_t}^{-1} \mathbf{A}_{s_{t}} \\
\boldsymbol{\theta}_{t-1 \mid t-1}^{b}=& \mathbf{C}_{s_{t-1}}^{T} \mathbf{S}_{s_{t_1}}^{-1} \boldsymbol{y}_{t-1}+\mathbb{I}[d_{t-1} = 1](\mathbf{R}_{s_{t}}^{D T}\boldsymbol{\kappa}^D_{t} - \mathbf{R}_{s_{t}}^D \boldsymbol{\Omega}^{D}_{t-1} \boldsymbol{r}_{s_{t}}^D + \mathbf{R}_{s_{t}}^{S T}\boldsymbol{\kappa}^S_{t} - \mathbf{R}_{s_{t-1}}^S \boldsymbol{\Omega}^{S}_{t-1} \boldsymbol{r}_{s_{t-1}}^S) -\mathbf{A}_{s_t}^{T} \mathbf{Q}_{s_t}^{-1} \mathbf{a}_{s_t} \nonumber \\
&+\mathbf{A}_{s_t}^{T} \mathbf{Q}_{s_t}^{-1}(\mathbf{Q}_{s_t}^{-1}+\mathbf{C}_{s_t}^{T} \mathbf{S}_{s_{t}}^{-1} \mathbf{C}_{s_t}+\boldsymbol{\Lambda}_{t+1, t})^{-1}(\boldsymbol{\theta}_{t+1, t} + \mathbf{C}_{s_t}^{T} \mathbf{S}_{s_{t}}^{-1} \boldsymbol{y}_{t} - \mathbf{C}_{s_t}^{T} \mathbf{S}_{s_{t}}^{-1}\boldsymbol{c}_{s_t} +\mathbf{Q}_{s_t}^{-1} \mathbf{a}_{s_t}) \\
=&\mathbf{C}_{s_{t-1}}^{T} \mathbf{S}_{s_{t_1}}^{-1} \boldsymbol{y}_{t-1}+\mathbb{I}[d_{t-1} = 1](\mathbf{R}_{s_{t}}^{D T}\boldsymbol{\kappa}^D_{t} - \mathbf{R}_{s_{t}}^D \boldsymbol{\Omega}^{D}_{t-1} \boldsymbol{r}_{s_{t}}^D + \mathbf{R}_{s_{t-1}}^{S T}\boldsymbol{\kappa}^S_{t} - \mathbf{R}_{s_{t-1}}^S \boldsymbol{\Omega}^{S}_{t-1} \boldsymbol{r}_{s_{t-1}}^S) -\mathbf{A}_{s_t}^{T} \mathbf{Q}_{s_t}^{-1} \mathbf{a}_{s_t} \nonumber \\
&+\mathbf{A}_{s_t}^{T} \mathbf{Q}_{s_t}^{-1}(\mathbf{Q}_{s_t}^{-1}+\boldsymbol{\Lambda}_{t \mid t})^{-1}(\boldsymbol{\theta}_{t \mid t}^{b} +\mathbf{Q}_{s_t}^{-1} \mathbf{a}_{s_t})
\end{aligned}
\end{equation}
We initialize at time T with
\begin{equation}
\begin{aligned}
\boldsymbol{\Lambda}_{T \mid T}^{b} &=\mathbf{C}_{s_T}^{T} \mathbf{S}_{s_{T}}^{-1} \mathbf{C}_{s_T} \\
\boldsymbol{\theta}_{T \mid T}^{b} &=\mathbf{C}_{s_T}^{T} \mathbf{S}_{s_{T}}^{-1} \boldsymbol{y}_{T}- \mathbf{C}_{s_T}^{T} \mathbf{S}_{s_{T}}^{-1}\boldsymbol{c}_{s_t}
\end{aligned}
\end{equation}
After computing the messages $m_{t+1, t}\left(\boldsymbol{x}_{t}\right)$ backwards in time, we sample the state sequence $\boldsymbol{x}_{1: T}$ working forwards in time. As with the discrete mode sequence, one can decompose the posterior distribution of the state sequence as
\begin{equation}
\begin{aligned}
p\left(\boldsymbol{x}_{1: T} \mid \boldsymbol{y}_{1: T}, s_{1: T}, \boldsymbol{\theta}\right) &=p\left(\boldsymbol{x}_{T} \mid \boldsymbol{x}_{T-1}, \boldsymbol{y}_{1: T}, s_{1: T}, \boldsymbol{\theta}\right) p\left(\boldsymbol{x}_{T-1} \mid \boldsymbol{x}_{T-2}, \boldsymbol{y}_{1: T}, s_{1: T}, \boldsymbol{\theta}\right) \\
& \cdots p\left(\boldsymbol{x}_{2} \mid \boldsymbol{x}_{1}, \boldsymbol{y}_{1: T}, s_{1: T}, \boldsymbol{\theta}\right) p\left(\boldsymbol{x}_{1} \mid \boldsymbol{y}_{1: T}, s_{1: T}, \boldsymbol{\theta}\right) .
\end{aligned}
\end{equation}
\begin{equation}
p\left(\boldsymbol{x}_{t} \mid \boldsymbol{x}_{t-1}, \boldsymbol{y}_{1: T}, s_{1: T}, \boldsymbol{\theta}\right) \propto p\left(\boldsymbol{x}_{t} \mid \boldsymbol{x}_{t-1}, \mathbf{A}_{s_{t}}, \mathbf{Q}_{s_t}\right) p\left(\boldsymbol{y}_{t} \mid \boldsymbol{x}_{t}, \mathbf{S}_{s_{t}}\right) m_{t+1, t}\left(\boldsymbol{x}_{t}\right)
\end{equation}

\begin{equation}
\begin{aligned}
p\left(\boldsymbol{x}_{t} \mid \boldsymbol{x}_{t-1}, \boldsymbol{y}_{1: T}, s_{1: T}, \boldsymbol{\theta}\right) & \propto \mathcal{N}\left(\boldsymbol{x}_{t} ; \mathbf{A}_{s_{t}} \boldsymbol{x}_{t-1} + \mathbf{a}_{s_t}, \mathbf{Q}_{s_t}\right) \mathcal{N}\left(\boldsymbol{y}_{t} ; \mathbf{C}_{s_t} \boldsymbol{x}_{t} + \mathbf{d}_{s_t}, \mathbf{S}_{s_{t}}\right) m_{t+1, t}\left(\boldsymbol{x}_{t}\right) \\
& \propto \mathcal{N}\left(\boldsymbol{x}_{t} ; \mathbf{A}_{s_{t}} \boldsymbol{x}_{t-1} + \mathbf{a}_{s_t}, \mathbf{Q}_{s_t}\right) \mathcal{N}^{-1}\left(\boldsymbol{x}_{t} ; \boldsymbol{\theta}_{t \mid t}^{b}, \boldsymbol{\Lambda}_{t \mid t}^{b}\right) \\
& \propto \mathcal{N}^{-1}\left(\boldsymbol{x}_{t} ; \mathbf{Q}^{-1}_{s_{t}}\left(\mathbf{A}_{s_{t}} \boldsymbol{x}_{t-1} + \mathbf{a}_{s_t}\right)+\boldsymbol{\theta}_{t \mid t}^{b}, \mathbf{Q}^{-1}_{s_{t}}+\boldsymbol{\Lambda}_{t \mid t}^{b}\right)
\end{aligned}
\end{equation}

\section{Full learning pseudocode}\label{section:pseudocodes}

To make it easier to understand the whole learning and inference procedure, we present pseudocode. Algorithm \ref{alg:x_sample} and Algorithm \ref{alg:stablefilterBackward} are algorithms adapted from \cite{fox_bayesian_2011} to have a common notation with our paper.

Let us define $z_t := (s_t, d_t)$.
Let's take note that we compute forward  probability as
\begin{equation}
\label{eqn:forward}
\begin{aligned}
\alpha_{t}\left(z_{t}\right) &=p\left(z_{t}, \mathbf{y}_{1: t}, \boldsymbol{x}_{1: t}\right)\\ 
&=\sum_{z_{t-1}}p\left(z_{t} \mid z_{t-1}\right)  p\left(\mathbf{y}_{t}, \mathbf{x}_{t} \mid z_{t}\right) \alpha_{t-1}\left(z_{t-1}\right).
\end{aligned}
\end{equation}

The procedure of sampling pseudo-observations is presented in the Algorithm \ref{alg:x_sample}. It uses the messages computed using \ref{alg:stablefilterBackward}.
The complete model fitting procedure is given by the Algorithm \ref{alg:beam}.
\begin{figure*}[h]
\begin{algorithm}[H]
    \caption{Fit the REDSLDS}
    \label{alg:beam}
    \begin{algorithmic}
%\STATE \textbf{Initialise:}
% \STATE Initialise parameters $\boldsymbol{A}$, $\boldsymbol{\theta}.$ Initialize $u_t$ small $\forall\, t$
%\STATE set $\mathcal{U}$ to a small value for all $u_t \in \mathcal{U}$
% \STATE Initialise $z_{1:T}$ randomly
% \STATE Initialise $x_{1:T}$ using Kalman filter
\FOR{iteration $ \in \{1,2,3,\ldots \}$}
   % \STATE \textbf{Forwards pass}:
    \STATE \textbf{Forward}: compute \eqref{eqn:forward} dynamically to get $\alpha_t(s_t)$ given, $x_{1:T}$ and $y_{1:T}$
    %\STATE \textbf{Backwards sample}:
    \STATE \textbf{Backward}: sample $z_T \sim \alpha_T(z_T)$
    \FOR{$t \in \{T, T-1, \ldots, 1\}$}
        \STATE sample $z_{t-1} \sim p(z_{t} | z_{t-1})\alpha_{t-1}(z_{t-1})$
    \ENDFOR
\STATE sample pseudoobservations $x_{1:T}$ using Algorithm \ref{alg:x_sample}
\STATE sample auxilirary variables $\omega^D_{1:T}, \omega^S_{1:T}$ using \eqref{eq:pq_posterior}
\STATE sample parameters $\boldsymbol{\theta}$ 
\ENDFOR
\end{algorithmic}
\end{algorithm}
 \hfill
\end{figure*}
\begin{figure*}[h]
\begin{algorithm}[H]
    \caption{Sample pseudobservations}
    \label{alg:x_sample}
    \begin{algorithmic}
    \STATE For each $t \in\{T, \ldots, 1\}$, recursively compute $\left\{\boldsymbol{\theta}_{t \mid t}^{b}, \boldsymbol{\Lambda}_{t \mid t}^{b}\right\}$ as in Algorithm \ref{alg:stablefilterBackward}.
    \STATE Working sequentially forward in time sample
    $$
    \mathcal{N}^{-1}\left(\boldsymbol{x}_{t} ; \mathbf{Q}^{-1}_{s_{t}}\left(\mathbf{A}_{s_{t}} \boldsymbol{x}_{t-1} + \mathbf{a}_{s_t}\right)+\boldsymbol{\theta}_{t \mid t}^{b}, \mathbf{Q}^{-1}_{s_{t}}+\boldsymbol{\Lambda}_{t \mid t}^{b}\right) .
    $$
\end{algorithmic}
\end{algorithm}
 \hfill
\end{figure*}
\begin{figure*}[h]
\begin{algorithm}[H]\hspace*{-6pt}
\begin{flushleft}
\begin{enumerate}
\item Initialize filter with
\begin{align*}
\boldsymbol{\Lambda}_{T \mid T}^{b} &=\mathbf{C}_{s_T}^{T} \mathbf{S}_{s_{T}}^{-1} \mathbf{C}_{s_T} \\
\boldsymbol{\theta}_{T \mid T}^{b} &=\mathbf{C}_{s_T}^{T} \mathbf{S}_{s_{T}}^{-1} \boldsymbol{y}_{T}- \mathbf{C}_{s_T}^{T} \mathbf{S}_{s_{T}}^{-1}\boldsymbol{c}_{s_t}
\end{align*}
\item Working backwards in time, for each $t \in \{T-1,\dots,1\}$:
\begin{enumerate}
\item Compute
\begin{align*}
\tilde{\boldsymbol{J}}_{t+1} &= \boldsymbol{\Lambda}^{b}_{t+1|t+1}(\boldsymbol{\Lambda}^{b}_{t+1|t+1} + \boldsymbol{\Sigma}^{-(s_{t+1})})^{-1}\\
\tilde{\boldsymbol{L}}_{t+1} &= \boldsymbol{I} - \tilde{\boldsymbol{J}}_{t+1}.
\end{align*}
\item Predict
\begin{align*}
\boldsymbol{\Lambda}_{t+1,t} &=
\boldsymbol{A}_{s_{t+1}}^T(\tilde{\boldsymbol{L}}_{t+1}\boldsymbol{\Lambda}^b_{t+1|t+1}\tilde{\boldsymbol{L}}_{t+1}^T
+
\tilde{\boldsymbol{J}}_{t+1}\boldsymbol{\Sigma}^{-(s_{t+1})}\tilde{\boldsymbol{J}}_{t+1}^T)\boldsymbol{A}_{s_{t+1}}\\
&+ \mathbb{I}[d_{t} = 1](\boldsymbol{\Omega}^{S}_{t} \mathbf{R}_{s_{t}}^{S T}\mathbf{R}_{s_{t}}^S + \boldsymbol{\Omega}^{D}_{t} \mathbf{R}_{s_{t+1}}^{D T}\mathbf{R}_{s_{t+1}}^D) \\
\boldsymbol{\theta}_{t+1,t} &=
\boldsymbol{A}_{s_{t+1}}^T\tilde{\boldsymbol{L}}_{t+1}(\boldsymbol{\theta}^b_{t+1|t+1}-\boldsymbol{\Lambda}_{t+1|t+1}^b\boldsymbol{a}_{s_{t+1}}) \\
&+\mathbb{I}[d_{t} = 1](\mathbf{R}_{s_{t+1}}^{D T}\boldsymbol{\kappa}^D_{t+1} - \mathbf{R}_{s_{t+1}}^D \boldsymbol{\Omega}^{D}_{t} \boldsymbol{r}_{s_{t+1}}^D + \mathbf{R}_{s_{t}}^{S T}\boldsymbol{\kappa}^S_{t+1} - \mathbf{R}_{s_{t}}^S \boldsymbol{\Omega}^{S}_{t} \boldsymbol{r}_{s_{t}}^S)
\end{align*}
\item Update
\begin{align*}
\boldsymbol{\Lambda}^b_{t|t} &= \boldsymbol{\Lambda}_{t+1,t} +
\boldsymbol{C}_{s_{t}}^T\boldsymbol{S}_{s_{t}}^{-1}\boldsymbol{C}_{s_{t}}\\
\boldsymbol{\theta}^b_{t|t} &= \mathbf{C}_{s_t}^{T} \mathbf{S}_{s_{t}}^{-1} \boldsymbol{y}_{t} - \mathbf{C}_{s_t}^{T} \mathbf{S}_{s_{t}}^{-1}\boldsymbol{c}_{s_t}+\boldsymbol{\theta}_{t+1, t}
\end{align*}
\end{enumerate}
\item Set
\begin{align*}
\boldsymbol{\Lambda}^b_{0|0} &= \boldsymbol{\Lambda}_{1,0}\\
\boldsymbol{\theta}^b_{0|0} &= \boldsymbol{\theta}_{1,0}
\end{align*}
\end{enumerate}
\end{flushleft}
%\begin{singlespace}
\caption{The backwards Kalman information
filter.} \label{alg:stablefilterBackward}
%\end{singlespace}
\end{algorithm}
\end{figure*}

\section{Extended experiment discussion}\label{section:extendedresults}

To assess our model's performance under conditions with limited independent observations, we tested it on three distinct benchmarks. These benchmarks included a simulated dataset, NASCAR$^{\circledR}$ (simulated by \citet{linderman_bayesian_2017}), and two publicly available real-world datasets.

The first real-world dataset is based on honeybee dance patterns, and the second records mouse behavior. In all experiments, labels were obtained in a \emph{fully self-supervised manner}. 

NASCAR$^{\circledR}$ \citep{linderman_bayesian_2017} serves as a controlled setting to illustrate the model's competence in capturing and segmenting dynamic behaviors into distinct states, facilitating a comparison with traditional rSLDS models and demonstrating REDSLDS's superior segmentation abilities. As mentioned in the main paper, to increase the challenge, the data were sampled in a manner that mimics taking independent samples without clear insights into process periodicity.

The honeybee dance dataset \citep{oh_learning_2008} was selected for its unique behavioral patterns, where honeybees convey food locations through complex dance movements. This dataset poses a challenge in accurately segmenting the rapid and unstable movements typical of the waggle phase of the dance. It aligns with the assumption that the data consist of independent trials from a common process, making it ideal for testing the model's effectiveness in real-world scenarios with distinct patterned behaviors.

The mouse behavior dataset from the BehaveNet study \citep{batty_behavenet_2019} is particularly demanding due to its high dimensionality and the intricate behavioral patterns observed during tasks. Selecting this dataset allows examination of REDSLDS's ability to segment and interpret complex behavioral data, underscoring its relevance for high-dimensional data scenarios.

We have selected three notable benchmark datasets for our analysis. These datasets are prevalently utilized within the research community and have been employed in various studies, including those by \citet{linderman_bayesian_2017, nassar_tree-structured_2019, ansari_deep_2021, lee_switching_2023, hutter_disentangled_2021}. For performance baselines in our study, we include the SLDS, rSLDS, and EDSLDS models. It is important to highlight that the EDSLDS model diverges from traditional approaches by substituting $Dur_{s_t, x_{t-1}}$ with a categorical distribution specifically tailored for the duration variable.

\paragraph{Initialization} Establishing effective initial conditions for SLDS-based models remains an active area of investigation, with numerous methodologies currently under study \citep{linderman_bayesian_2017, nassar_tree-structured_2019, slupinski_improving_2024}. Our approach starts by initializing the latent continuous state through principal component analysis (PCA), in accordance with the strategy employed by \citet{linderman_bayesian_2017}. Subsequently, five Autoregressive Hidden Markov Models (ARHMMs) are fitted, and the model that maximizes the log-likelihood is selected. The state sequences decoded from this model provide the initial state assignments for the REDSLDS model. It was found that model convergence accelerated when, following ARHMM initialization, an additional five iterations were conducted with the model parameters held constant at $z_{1:T}$ and $\mathbf{x}_{1:T}$. This phase is referred to as “Init I” in Table \ref{tab:nascar:results}. Conversely, in more complex scenarios, the REDSLDS model exhibited better performance when these supplementary iterations were omitted, a condition we term “Init II”. This trend was not markedly observed in simpler models such as rSLDS, likely due to the inherent complexity of the model that necessitates an extensive exploration of the parameter space during the initial stages.

Detailed priors are provided in Section \ref{section:priors} for each run as well as the sampling scheme.
% \begin{figure}[!htb]
% \centering
% \includegraphics[width=0.3\linewidth]{Figures/nascar_trajectory.pdf}
% \caption{Trajectory of \emph{NASCAR$^{\circledR}$} used to test the models.}
% \label{fig:nascar:trajectory}
% 
% \end{figure}
% 
% \subsection{NASCAR$^{\circledR}$}\label{section:nascar}
% 

We commence by presenting a straightforward example, wherein the dynamics generate elliptical patterns akin to those observed on a NASCAR\textsuperscript{\circledR} track, as depicted in Figure \ref{fig:nascar:trajectory} \citep{linderman_bayesian_2017}. In this scenario, the dynamics are governed by four discrete states, denoted by $s_t \in \{1, \ldots, 4\}$, which in turn control a continuous latent state represented in two-dimensional space, $\mathbf{x}_t \in \mathbb{R}^2$. The observations, $\mathbf{y}_t \in \mathbb{R}^{10}$, are derived by linearly mapping the latent state and perturbing it with Gaussian noise. Various studies have documented the efficiency of rSLDS models when handling such data \citep{linderman_bayesian_2017, nassar_tree-structured_2019}.

Our focus, however, is to scrutinize the efficacy of SLDS-based models when they are applied to independent sequences that originate from an identical distribution. To this end, we generated 10 independent NASCAR\textsuperscript{\circledR} runs, each consisting of 12,000 sample points. Each run was then partitioned into $S \in \{5, 10, 15, 20\}$ segments (the multiples of chunks were arbitrarily chosen; they could have been $ \{3, 6, 12, 24\}$ or $\{2, 5, 12, 20\}$ as well). From these segments, we randomly sampled $N = 0.8S$ segments, resulting in an aggregate dataset comprising of 9,600 points. Each model underwent 10,000 iterations of Gibbs sampling.

The metrics assessing segmentation quality alongside the log-likelihood scores for the models are summarized in Table \ref{tab:nascar:results}. Sample segmentations derived from rSLDS and REDSLDS models are illustrated in Figures \ref{fig:nascar_models_ll} and \ref{fig:nascar:split:segmentation}. An intriguing observation is that SLDS models frequently register the highest log-likelihood scores, although they do not always achieve the top segmentation performance. It is imperative to note that model selection should not rely solely on the log-likelihood values, particularly when latent Markov states are mixed with Gaussian distributions. This is because of potential singularities in the likelihood function, where a Gaussian component can collapse onto a specific data point as discussed by \citet{bishop_pattern_2006}.

Upon examining the results more comprehensively, it is evident that our model generally secured the highest segmentation scores, with the notable exception of the scenario involving 10 splits, where the EDSLDS model distinctly outperformed others. Importantly, as previously highlighted during the initialization discussion, SLDS-based models exhibit considerable sensitivity to initialization conditions. This sensitivity suggests that EDSLDS's superior performance in this particular case might have been due to particularly favorable initialization conditions.

Our analysis indicates that, for most models, increasing the number of splits generally leads to a reduction in the log-likelihood of the model. One notable exception is the case with five splits in the "Init I" model, which appears to be underfitted according to our observations. This trend suggests that the model achieves a better fit to the data relative to other models. This observation is consistent with our hypothesis that shorter sequences present a more significant challenge to all models in terms of accurate data representation. The recurrent explicit duration variables compel the model to remain in a given state for a more extended period, whereas in (r)SLDS, state transitions can occur more freely. As illustrated in Table \ref{tab:nascar:results}, we gain advantages from integrating both recurrence and duration modeling into the approach.

\subsection{Dancing Bee}\label{section:bee}
We leveraged the publicly available dancing bee dataset \citep{oh_learning_2008}, which is renowned for its intricate and complex characteristics and has been previously scrutinized in the context of time series segmentation. This dataset encompasses detailed information regarding the movements of six individual honeybees as they execute the waggle dance. Specifically, it includes the bees' coordinates within a two-dimensional plane, as well as the angles at which they are oriented at each time step.

The communication of food sources among honeybees is facilitated through a series of dances that occur within the hive. These dances incorporate particular movement patterns, such as the waggle, right turn, and left turn. The waggle dance, for instance, involves a bee traversing in a straight line while vigorously shaking its body from side to side. In contrast, the turning dances are characterized by the bee rotating either clockwise or counterclockwise. The data recorded for this analysis is represented as $\boldsymbol{y}_t=\left[\begin{array}{llll}\cos \left(\theta_t\right), & \sin \left(\theta_t\right), & x_t, & y_t\end{array}\right]^T$, with $\left(x_t, y_t\right)$ depicting the two-dimensional spatial coordinates of the bee's body and $\theta_t$ denoting the orientation angle of its head.

Given that the dataset includes distinct recordings of individual bee dances, it conforms to our presumption that the data consists of independent trials derived from a common underlying process. To evaluate our models, we scored them on the entire dataset, as well as on a subset consisting of the last three subsequences. This choice was made because the initial three subsequences exhibit slightly different properties and varying lengths. Each model underwent sampling through 10,000 iterations of the Gibbs sampling algorithm.

The Table \ref{tab:nascar:results} displays the complete segmentation scores and model log-likelihoods. It can be seen that our model outperforms all of the other baselines. For comparison, REDSDS for \citet{ansari_deep_2021} scored the precision of ($0.73\pm0.10$). Figure \ref{fig:bee:segmentation} presents the sample segmentations.

The rSLDS algorithm encounters significant difficulties in accurately modeling long-term motion patterns, which consequently leads to the creation of a multitude of segments. This issue becomes particularly pronounced during the ``waggle'' phase observed in bee dances, a phase distinguished by swift and erratic movements, as depicted in Figure \ref{fig:bee:rSLDS}.
\raggedbottom
\subsection{BehaveNet}\label{section:behavenet}

Additionally, we applied our model to a publicly available dataset containing mouse behavior data \citep{batty_behavenet_2019}. In this specific experiment, a mouse, whose head was immobilized, engaged in a visual decision-making task while neural activity within the dorsal cortex was monitored using wide-field calcium imaging. Our focus was exclusively on the behavior video data, which was recorded in gray-scale at a resolution of 128x128 pixels per frame. The behavior was captured using two cameras, one offering a lateral view and the other a ventral view. Owing to the high dimensional nature of the behavior video data, we directly utilized dimension reduction findings from earlier research. These findings comprised of 9-dimensional continuous variables estimated through the use of a convolutional autoencoder.

\emph{Since the dataset lacks labels, we approached the analysis as a qualitative experiment.} Each model was sampled over 5000 Gibbs iterations. In this experiment, we utilized only the rSLDS and REDSLDS models, employing initialization scheme 2. From the comprehensive dataset, we randomly selected ten sequences with uniform probability and standardized the measurements appropriately.

% \begin{figure}[!htb]
% \centering
% \includegraphics[width=0.9\linewidth]{Figures/rSLDS_behavenet.pdf}
% \caption{The segmentation of  BehaveNet data obtained by rSLDS model. The figure clearly illustrates that the majority of the data clusters together in a single state.}
% \label{fig:behavenet:rSLDS}
% 

% \end{figure}

% \begin{figure}[!htb]
% \centering
% \includegraphics[width=0.9\linewidth]{Figures/redslds_behavenet.pdf}
% \caption{The segmentation of  BehaveNet data obtained by REDSLDS model. The presence of a notable degree of self-persistence among the states is apparent. The states exhibit different magnitudes, and the orange state, in particular, demonstrates the least variation in measurement.}
% 
% \label{fig:behavenet:redslds}
% \end{figure}

The application of the ARHMM model for segmenting this particular data set is prone to the issue of over-segmentation. This challenge has been recently addressed by \citep{costacurta_distinguishing_2022}. In their research, the authors introduced auxiliary variables that incorporated additional noise into the traditional ARHMM framework. To further elaborate, in models that are built upon the SLDS architecture, additional noise is injected into the autoregressive process via the emission layer. Figure \ref{fig:behavenet:segmentation} illustrates the segmentation results produced by the rSLDS model, which attained the highest log-likelihood value of $-13420.80$. The figure clearly shows that the majority of the data converges into a single state. Moving on to the REDSLDS model, its segmentation results are depicted in Figure \ref{fig:behavenet:redslds}, boasting a log-likelihood value of $-13245.71$. This figure vividly highlights a pronounced self-persistence among the states. The different states demonstrate distinct amplitude levels, with the orange state, in particular, showing the minimal variation in measurement.
\pagebreak
\section{Prior details}
\label{section:priors}

For every experiment we set multivariate Gaussian prior on columns of $R^S$ matrix, with mean \textbf{0} and covariance $\mathbf{\Sigma}^{R^S}_0$. Similarly, we set a multivariate Gaussian prior in columns of the matrix $R^D$, with mean \textbf{ 0} and covariance~$\boldsymbol{\Sigma}^{R^D}_0$.

We utilized the Matrix Normal inverse-Wishart ($\operatorname{MNIW}(\mathbf{M}, \mathbf{V}, \mathbf{S}, n)$) prior to use autoregressive dynamics and emissions in each of the experiments. A detailed explanation of the MNIW prior can be found in \citep{fox_bayesian_2011}.

For both emission and duration, we set $\mathbf{M}_0$ to $\mathbf{0}$, $n_0=N+2$ where $N$ is the dimensionality of the observations, and $\mathbf{S}_{0} = 0.075\cdot0.75 \cdot \Bar{S}$, where $\Bar{S}$ is the empirical covariance matrix.

For both emission and duration, we set $\mathbf{M}_0$ to $\mathbf{0}$, $n_0=M+2$ where $M$ is the dimensionality of the latent space and $\mathbf{S}_{0} = 0.75\cdot0.75 \cdot \Bar{S}$, where $\Bar{S_{PCA}}$ is the empirical covariance matrix of the PCA transformed observations.

By $\mathbf{I}$ we denote identity matrix.

The rest of the priors are described in the tables below.

We chose the set of parameters by performing a grid search and selecting parameterization maximizing weighted F1.

\subsection{Dancing Bee}
\begin{table}[H]
\centering
\caption{Priors used in the dancing bee experiment presented in Section 4.2}
\label{tab:bees:results}
\begin{tabular}{lrrrrrr}
\toprule
\multicolumn{7}{c}{Sequences 4, 5, 6} \\
\midrule
{} & SLDS & EDSLDS & rSLDS I & REDSLDS I& rSLDS II & REDSLDS II \\
\midrule
Latent dimension & 2 &  2 & 2 &  2 & 2 &  2 \\
Dynamics $\boldsymbol{V}_0$ & $\mathbf{I}$ & 0.1$\mathbf{I}$ & $\mathbf{I}$ & $\mathbf{I}$ & 0.1$\mathbf{I}$ & $\mathbf{I}$ \\
Emission $\boldsymbol{V}_0$ & $\mathbf{I}$ & 0.1$\mathbf{I}$ & $0.1\mathbf{I} $& $0.1\mathbf{I}$& $\mathbf{I}$ & 0.1$\mathbf{I}$ \\
$\boldsymbol{\Sigma}^{R^S}_0$ & - & - & $10000\mathbf{I}$ & $\mathbf{I}$ & 0.0001$\mathbf{I}$ & 0.0001$\mathbf{I}$ \\
$\boldsymbol{\Sigma}^{R^D}_0$ & - & - & - & $10000\mathbf{I}$ & - & 10000$\mathbf{I}$ \\
\midrule
\multicolumn{7}{c}{All sequences} \\
\midrule
{} & SLDS & EDSLDS & rSLDS I & REDSLDS I& rSLDS II & REDSLDS II \\
\midrule
Latent dimension & 2 &  2 & 2 &  2 & 2 &  2 \\
Dynamics $\boldsymbol{V}_0$ & $\mathbf{I}$ & 0.1$\mathbf{I}$ & $0.1\mathbf{I}$& $\mathbf{I}$ & $\mathbf{I}$ & 0.$\mathbf{I}$ \\
Emission $\boldsymbol{V}_0$ & $\mathbf{I}$ & 0.1$\mathbf{I}$ & $\mathbf{I}$ & $0.1\mathbf{I} $ & 0.1$\mathbf{I}$ & $\mathbf{I}$ \\
$\boldsymbol{\Sigma}^{R^S}_0$ & - & - & $0.0001\mathbf{I}$ & $10000\mathbf{I}$ & 0.0001$\mathbf{I}$ & $\mathbf{I}$ \\
$\boldsymbol{\Sigma}^{R^D}_0$ & - & - & - & $10000\mathbf{I}$ & - & 10000$\mathbf{I}$ \\
\bottomrule
\end{tabular}
\end{table}

\subsection{Nascar}
\begin{table}[H]
\centering
\caption{Priors used in the NASCAR$^{\circledR}$ experiment presented in Section 4.1}
\label{tab:bees:results}
\begin{tabular}{lrrrrrr}
\toprule
\multicolumn{7}{c}{Split 5} \\
\midrule
{} & SLDS & EDSLDS & rSLDS I & REDSLDS I& rSLDS II & REDSLDS II \\
\midrule
Latent dimension & 2 &  2 & 2 &  2 & 2 &  2 \\
Dynamics $\boldsymbol{V}_0$ & $\mathbf{I}$ & $\mathbf{I}$ & $\mathbf{I}$ & $0.1\mathbf{I}$ & 0.1$\mathbf{I}$ & $\mathbf{I}$ \\
Emission $\boldsymbol{V}_0$ & $\mathbf{I}$ & $\mathbf{I}$ & $0.1\mathbf{I}$ & $\mathbf{I}$ & 0.1$\mathbf{I}$ & $\mathbf{I}$ \\
$\boldsymbol{\Sigma}^{R^S}_0$ & - & - & $0.0001\mathbf{I}$ & $\mathbf{I}$ & 10000$\mathbf{I}$ & 0.000$\mathbf{I}$ \\
$\boldsymbol{\Sigma}^{R^D}_0$ & - & - & - & $10000\mathbf{I}$ & - & 10000$\mathbf{I}$ \\
\midrule
\multicolumn{7}{c}{Split 10} \\
\midrule
{} & SLDS & EDSLDS & rSLDS I & REDSLDS I& rSLDS II & REDSLDS II \\
\midrule
Latent dimension & 2 &  2 & 2 &  2 & 2 &  2 \\
Dynamics $\boldsymbol{V}_0$ & $\mathbf{I}$ & $\mathbf{I}$ & $\mathbf{I}$ & $0.1\mathbf{I}$ & $\mathbf{I}$ & $\mathbf{I}$ \\
Emission $\boldsymbol{V}_0$ & $\mathbf{I}$ & $\mathbf{I}$ & $\mathbf{I}$ & $0.1\mathbf{I}$ & $\mathbf{I}$ & $\mathbf{I}$ \\
$\boldsymbol{\Sigma}^{R^S}_0$ & - & - & $0.0001\mathbf{I}$ & $0.0001\mathbf{I}$ & $\mathbf{I}$ & 10000$\mathbf{I}$ \\
$\boldsymbol{\Sigma}^{R^D}_0$ & - & - & - & $10000\mathbf{I}$ & - & 10000$\mathbf{I}$ \\
\midrule
\multicolumn{7}{c}{Split 15} \\
\midrule
{} & SLDS & EDSLDS & rSLDS I & REDSLDS I& rSLDS II & REDSLDS II \\
\midrule
Latent dimension & 2 &  2 & 2 &  2 & 2 &  2 \\
Dynamics $\boldsymbol{V}_0$ & $0.1\mathbf{I}$ & $0.0001\mathbf{I}$ & $\mathbf{I}$ & $0.1\mathbf{I}$ & $\mathbf{I}$ & 0.000$\mathbf{I}$ \\
Emission $\boldsymbol{V}_0$ & $0.1\mathbf{I}$ & 0.1$\mathbf{I}$ & $\mathbf{I}$ & $\mathbf{I}$ & 0.$\mathbf{I}$ & 0.000$\mathbf{I}$ \\
$\boldsymbol{\Sigma}^{R^S}_0$ & - & - & $0.0001\mathbf{I}$ & $\mathbf{I}$ & $\mathbf{I}$ & 0.000$\mathbf{I}$ \\
$\boldsymbol{\Sigma}^{R^D}_0$ & - & - & - & $10000\mathbf{I}$ & - & 10000$\mathbf{I}$ \\
\midrule
\multicolumn{7}{c}{Split 20} \\
\midrule
{} & SLDS & EDSLDS & rSLDS I & REDSLDS I& rSLDS II & REDSLDS II \\
\midrule
Latent dimension & 2 &  2 & 2 &  2 & 2 &  2 \\
Dynamics $\boldsymbol{V}_0$ & $\mathbf{I}$ & $\mathbf{I}$ & $0.1\mathbf{I}$ & $0.1\mathbf{I}$ & 0.$\mathbf{I}$ & $\mathbf{I}$ \\
Emission $\boldsymbol{V}_0$ & $\mathbf{I}$ & $0.1\mathbf{I}$ & $0.1\mathbf{I}$ & $\mathbf{I}$ & 0.1$\mathbf{I}$ & $\mathbf{I}$ \\
$\boldsymbol{\Sigma}^{R^S}_0$ & - & - & $10000\mathbf{I}$ & $10000\mathbf{I}$ & 0.0001$\mathbf{I}$ & 0.0001$\mathbf{I}$ \\
$\boldsymbol{\Sigma}^{R^S}_D$ & - & - & - & $10000\mathbf{I}$ & - & 10000$\mathbf{I}$ \\
\bottomrule
\end{tabular}
\end{table}

\subsection{BehaveNet}
\begin{table}[H]
\centering
\caption{Priors used in the mouse behavior experiment presented in Section 4.3}
\label{tab:bees:results}
\begin{tabular}{lrr}
\toprule
{} &  rSLDS & REDSLDS \\
\midrule
Latent dimension & 3 &  3  \\
Dynamics  & 0.1$\mathbf{I}$ &  0.1$\mathbf{I}$ \\
Emission & 0.1$\mathbf{I}$ & 0.1$\mathbf{I}$ \\
$\boldsymbol{\Sigma}^{R^S}_0$  & 0.0001$\mathbf{I}$ & $\mathbf{I}$ \\
$\boldsymbol{\Sigma}^{R^D}_0$ &  - & 10000$\mathbf{I}$ \\
\bottomrule
\end{tabular}
\end{table}

\begin{table*}[!p]
\centering
  \rotatebox{90}{% Rotate the minipage by the desired angle
    \scalebox{.8}{\begin{minipage}{1.6\textwidth}
\caption{Results of data segmentation. The means and standard deviations are calculated from ten separate MCMC runs that are independent. We measured \textbf{acc}uracy, \textbf{w}eighted \textbf{F1}, \textbf{m}acro \textbf{F1} and \textbf{l}og-\textbf{l}ikelihood. The best scores using Init 1 are \underline{underlined}. The best results overline are \textbf{bold}.}
\label{tab:nascar:results}
\begin{tabular}{lrrrrrr}
\toprule
\rowcolor{DarkGray}
\multicolumn{7}{c}{\emph{NASCAR$^{\circledR}$} dataset} \\
\midrule
\rowcolor{Gray}
\multicolumn{7}{c}{Split 5} \\
\midrule
\rowcolor{LightGray}
{} & SLDS (Init I) & EDSLDS (Init I) & rSLDS (Init I) & REDSLDS (Init I) & rSLDS (Init II) & REDSLDS (Init II) \\
\midrule
Acc      & $0.33 \pm 0.03$ & $0.34 \pm 0.01$  & $0.31 \pm 0.01$ & $\underline{0.63 \pm 0.13}$ & $0.54 \pm 0.08$ & $\mathbf{0.71 \pm 0.01}$ \\
Ll & $-8.93 \cdot 10^{12} \pm 4.34 \cdot 10^{12}$ & $-4.36 \cdot 10^{12} \pm 6.92 \cdot 10^{12}$ & $-1.16  \cdot 10^{12} \pm 5.07  \cdot 10^{11}$ & $\underline{-1.21 \cdot 10^{6} \pm 7.86 \cdot 10^{5}}$ & $7.84 \cdot 10^{4} \pm 1.39 \cdot 10^{4}$ & $\mathbf{8.07 \cdot 10^{4} \pm 2.30 \cdot 10^{3}}$ \\
% Ll & $-8934299523848.10 \pm 4344304550615.47$ & $-4362622021513.77 \pm 6920294609526.43$ & $-1159533454419.30 \pm 506710924807.61$ & $-1208971.66 \pm 786414.94$ & $78410.94 \pm 1386.08$ & $\mathbf{80704.46 \pm 2299.40}$ \\

M F1 & $0.22 \pm 0.06$ & $0.23 \pm 0.04$ & $0.24 \pm 0.02$ & $\underline{0.50 \pm 0.15}$ &$0.46 \pm 0.09$ & $\mathbf{0.58 \pm 0.02}$ \\
W F1 & $0.39 \pm 0.11$ & $0.39 \pm 0.06$ & $0.33 \pm 0.05$ & $\underline{0.71 \pm 0.08}$ & $0.60 \pm 0.05$ & $\mathbf{0.76 \pm 0.03}$ \\
\midrule
\rowcolor{Gray}
\multicolumn{7}{c}{Split 10} \\
\midrule
\rowcolor{LightGray}
{} & SLDS (Init I) & EDSLDS (Init I) & rSLDS (Init I) & REDSLDS (Init I) & rSLDS (Init II) & REDSLDS (Init II) \\
\midrule
Acc         & $0.65 \pm 0.00$ & $\underline{\mathbf{0.77 \pm 0.02}}$ & $0.48 \pm 0.02$ & $0.65 \pm 0.01$ & $0.51 \pm 0.01$ & $0.63 \pm 0.07$ \\
Ll & $\underline{\mathbf{9.44 \cdot 10^{4} \pm 1.28 \cdot 10^{3}}}$ & $8.61 \cdot 10^{4} \pm 1.28 \cdot 10^{4}$ & $9.13 \cdot 10^{4} \pm 1.01 \cdot 10^{3}$ & $9.04 \cdot 10^{4} \pm 1.90 \cdot 10^{3}$ & $7.04 \cdot 10^{4} \pm 5.81 \cdot 10^{2}$ & $7.23 \cdot 10^{4} \pm 2.15 \cdot 10^{3}$ \\
% Ll & $\mathbf{94430.44 \pm 1278.88}$ & $86142.95 \pm 12789.51$ & $91285.65 \pm 1013.15$ & $90427.19 \pm 1897.47$ & $70443.59 \pm 581.31$ & $72299.64 \pm 2149.22$ \\
M F1 & $0.53 \pm 0.00$ & $\underline{\mathbf{0.76 \pm 0.02}}$ & $0.41 \pm 0.02$ & $0.55 \pm 0.02$ & $0.47 \pm 0.01$ & $0.48 \pm 0.06$ \\
W F1 & $0.70 \pm 0.01$ & $\underline{\mathbf{0.76 \pm 0.02}}$ & $0.49 \pm 0.03$ & $0.68 \pm 0.02$ & $0.49 \pm 0.01$ & $0.68 \pm 0.07$ \\
\midrule
\rowcolor{Gray}
\multicolumn{7}{c}{Split 15} \\
\midrule
\rowcolor{LightGray}
{} & SLDS (Init I) & EDSLDS (Init I) & rSLDS (Init I) & REDSLDS (Init I) & rSLDS (Init II) & REDSLDS (Init II) \\
\midrule
Acc         & $0.50 \pm 0.12$ & $0.51 \pm 0.15$ & $\underline{0.52 \pm 0.05}$ & $\underline{0.52 \pm 0.09}$ & $0.48 \pm 0.08$ & $\mathbf{0.69 \pm 0.00}$ \\
Ll & $\underline{\mathbf{8.47 \cdot 10^{4} \pm 3.16 \cdot 10^{2}}}$ & $7.95 \cdot 10^{4} \pm 5.07 \cdot 10^{3}$ & $8.20 \cdot 10^{4} \pm 8.85 \cdot 10^{2}$ & $7.93 \cdot 10^{4} \pm 1.50 \cdot 10^{3}$ & $6.34 \cdot 10^{4} \pm 1.60  \cdot 10^{3}$ & $6.68  \cdot 10^{4} \pm 7.32 \cdot 10^{2}$ \\
% Ll & $\mathbf{84654.25 \pm 315.68}$ & $79541.20 \pm 5073.43$ & $82004.89 \pm 884.92$ & $79399.72 \pm 1499.90$ & $63352.30 \pm 1604.22$ & $66777.34 \pm 732.83$ \\
M F1 & $0.43 \pm 0.07$ & $\underline{0.49 \pm 0.14}$ & $0.42 \pm 0.05$ & $0.43 \pm 0.06$ & $0.35 \pm 0.10$ & $\mathbf{0.58 \pm 0.01}$ \\
W F1 & $0.51 \pm 0.14$ & $0.49 \pm 0.14$ & $\underline{0.55 \pm 0.05}$ & $\underline{0.55 \pm 0.09}$ & $0.54 \pm 0.06$ & $\mathbf{0.73 \pm 0.01}$ \\
\midrule
\rowcolor{Gray}
\multicolumn{7}{c}{Split 20} \\
\midrule
\rowcolor{LightGray}
{} & SLDS (Init I) & EDSLDS (Init I) & rSLDS (Init I) & REDSLDS (Init I) & rSLDS (Init II) & REDSLDS (Init II) \\
\midrule
Acc         & $0.40 \pm 0.06$ & $0.37 \pm 0.03$ & $0.35 \pm 0.01$ & $\underline{0.42 \pm 0.06}$ & $0.42 \pm 0.06$ & $\mathbf{0.51 \pm 0.05}$ \\
Ll & $\underline{\mathbf{4.69 \cdot 10^{4} \pm 2.73 \cdot 10^{3}}}$ & $4.52 \cdot 10^{4} \pm 4.00 \cdot 10^{3} $ & $4.91 \cdot 10^{4} \pm 6.44 \cdot 10^{3}$ & $4.62 \cdot 10^{4} \pm 3.61 \cdot 10^{3}$ &$3.43 \cdot 10^{4} \pm 4.80 \cdot 10^{3}$ & $3.83 \cdot 10^{4} \pm 2.64 \cdot 10^{3}$ \\
% Ll & $\mathbf{46904.49 \pm 2731.47}$ & $45224.87 \pm 3997.48 $ & $49090.33 \pm 6444.08$ & $46243.26 \pm 3612.68$ &$34330.58 \pm 4801.23$ & $38315.31 \pm 2640.15$ \\
M F1 & $0.21 \pm 0.08$ & $0.17 \pm 0.06$ & $0.18 \pm 0.04$ & $\underline{0.31 \pm 0.03}$ & $0.27 \pm 0.11$ & $\mathbf{0.41 \pm 0.05}$ \\
W F1 & $\underline{0.51 \pm 0.05}$ & $\underline{0.51 \pm 0.01}$ & $0.47 \pm 0.03$ & $0.46 \pm 0.10$ & $0.51 \pm 0.02$ & $\mathbf{0.54 \pm 0.05}$ \\
\midrule
\rowcolor{DarkGray}
\multicolumn{7}{c}{Bee Dataset} \\
\midrule
\rowcolor{Gray}
\multicolumn{7}{c}{Sequences 4, 5, 6} \\
\midrule
\rowcolor{LightGray}
{} & SLDS (Init I) & EDSLDS (Init I) & rSLDS (Init I) & REDSLDS (Init I) & rSLDS (Init II) & REDSLDS (Init II) \\
\midrule
Acc   &  $0.82 \pm 0.04$  &  $0.82 \pm 0.04$ & $0.37 \pm 0.01$ & $\underline{\mathbf{0.85 \pm  0.02}}$ & $0.43 \pm 0.03$ & $0.70 \pm 0.03$ \\
Ll & $-6.65 \cdot 10^{3} \pm 1.03 \cdot 10^{4}$ & $-6.63 \cdot 10^{3} \pm 1.03 \cdot 10^{4}$ & $-5.42 \cdot 10^{3} \pm 1.88 \cdot 10^{2}$ & $\underline{-4.73 \cdot 10^{3} \pm 2.99 \cdot 10^{2}}$ & $-3.19 \cdot 10^{8} \pm 3.50 \cdot 10^{8}$ & $\mathbf{-3.03 \cdot 10^{3} \pm 2.79 \cdot 10^{3}}$ \\
% Ll & $-6647.70 \pm 10314.93$ & $-6638.39 \pm 10314.93$ & $-5421.54 \pm 188.33$ & $-4738.56 \pm 298.81$ & $-318816084.49 \pm 350253515.68$ & $\mathbf{-3034.87 \pm 2794.29}$ \\
M F1 & $0.82 \pm 0.05$ & $0.82 \pm 0.04$ & $0.32 \pm 0.01$ & $\underline{\mathbf{0.86 \pm 0.02}}$ & $0.41 \pm 0.05$ & $0.69 \pm 0.03$ \\
W F1 & $0.82 \pm 0.04$ & $0.83 \pm 0.04$ & $0.40 \pm 0.00$ & $\underline{\mathbf{0.85 \pm 0.02}}$ & $0.45 \pm 0.03$ & $0.72 \pm 0.02$\\
\midrule
\rowcolor{Gray}
\multicolumn{7}{c}{All sequences} \\
\midrule
\rowcolor{LightGray}
{} & SLDS (Init I) & EDSLDS (Init I) & rSLDS (Init I) & REDSLDS (Init I) & rSLDS (Init II) & REDSLDS (Init II) \\
\midrule
Acc        & $0.39 \pm 0.02$& $0.41 \pm 0.00$ & $0.37 \pm 0.02$ & $\underline{0.42 \pm 0.02}$ & $0.40 \pm 0.00$ & $\mathbf{0.62 \pm 0.01}$ \\
Ll & $-2.82 \cdot 10^{7} \pm 3.98 \cdot 10^{7}$& $7.34 \cdot 10^{3} \pm 3.40 \cdot 10^{2}$& $-1.53  \cdot 10^{2} \pm 2.13 \cdot 10^{10}$ & $\underline{\mathbf{8.62 \cdot 10^{3} \pm 2.62 \cdot 10^{2}}}$ & $-3.80 \cdot 10^{3} \pm 1.27 \cdot 10^{2}$ & $-2.83 \cdot 10^{2} \pm 6.04 \cdot 10^{2}$ \\
% Ll & $-28157699.02 \pm 39836765.60$& $\mathbf{7345.82 \pm 3399.02}$& $-3797.54 \pm 127.11$ & $8623.59 \pm 2619.25$ & $-3797.54 \pm 127.11$ & $-2833.35 \pm 604.40$ \\
M F1 & $0.28 \pm 0.02$& $0.30 \pm 0.00$ & $0.30 \pm 0.02$ & $\underline{0.33 \pm 0.05}$ & $0.20 \pm 0.01$ & $\mathbf{0.62 \pm 0.01}$ \\
W F1 & $0.49 \pm 0.01$ & $ 0.50 \pm 0.00$ & $0.43 \pm 0.05$ & $\underline{0.50 \pm 0.00}$ & $0.57 \pm 0.00$ & $\mathbf{0.62 \pm 0.01}$ \\
\bottomrule
\end{tabular}
\end{minipage}}}
\end{table*}

\end{document}